\documentclass[preprint,10pt,authoryear]{elsarticle}

\usepackage[a4paper, margin=2cm]{geometry}
\usepackage{graphicx}
\usepackage{booktabs}
\usepackage{hyperref}
\usepackage{subcaption}
\usepackage{float}
\usepackage{placeins}
\usepackage{xparse}
\usepackage{csquotes}
\usepackage{xstring}
\usepackage{array}
\usepackage{multirow}
\usepackage{amsmath}
\usepackage{makecell}

\begin{document}
\begin{frontmatter}
\title{Cognitive networks reconstruct mindsets about STEM subjects and educational contexts in almost 1000 high-schoolers, University students and LLM-based digital twins}
\author[dipsco]{Francesco Gariboldi\texorpdfstring{\corref{equal}}{}}
\author[dipsco]{Emma Franchino\texorpdfstring{\corref{equal}}{}}
\author[dipsco]{Edith Haim}
\author[fisica]{Gianluca Lattanzi}
\author[bari]{Alessandro Grecucci}
\author[dipsco]{Massimo Stella}

\affiliation[dipsco]{
            organization={Department of Psychology and Cognitive Science, University of Trento},
            addressline={Corso Bettini, 31},
            city={Rovereto},
            postcode={38068},
            state={TN},
            country={Italy}}
\affiliation[bari]{
            organization ={Department of Psychology and Communication, University of Bari},
            addressline = {Via Scipione Crisanzio, 42},
            city ={Bari}, 
            postcode = {70122},
            state = {BA}, 
            country = {Italy}}
\affiliation[fisica]{
            organization={Department of Physics, University of Trento},
            addressline={Via Sommarive, 14},
            city={Povo},
            postcode={38123},
            state={TN},
            country={Italy}}

\cortext[equal]{The authors contributed equally.}

\begin{abstract}
Attitudes toward STEM develop from the interaction of conceptual knowledge, educational experiences, and affect. Here we use cognitive network science to reconstruct group mindsets as \textit{behavioural forma mentis networks} (BFMNs). In this case, nodes are cue words and free associations, edges are empirical associative links, and each concept is annotated with perceived valence. We analyse BFMNs from $N=994$ observations spanning high school students, university students, and early-career STEM experts, alongside LLM (GPT-oss) "digital twins" prompted to emulate comparable profiles. Focusing also on semantic neighbourhoods ("frames") around key target concepts (e.g., STEM subjects or educational actors/places), we quantify frames in terms of valence auras, emotional profiles, network overlap (Jaccard similarity), and concreteness relative to null baselines. Across student groups, \textit{science} and \textit{research} are consistently framed positively, while their core quantitative subjects (mathematics and statistics) exhibit more negative and anxiety related auras, amplified in higher math-anxiety subgroups, evidencing a STEM-science cognitive and emotional dissonance. High-anxiety frames are also less concrete than chance, suggesting more abstract and decontextualised representations of threatening quantitative domains. Human networks show greater overlapping between \textit{mathematics} and \textit{anxiety} than GPT-oss. The results highlight how BFMNs capture cognitive-affective signatures of mindsets towards the target domains and indicate that LLM-based digital twins approximate cultural attitudes but miss key context-sensitive, experience-based components relevant to replicate human educational anxiety.
\end{abstract}


\end{frontmatter}

\section{Introduction} \label{introduction}
Perceptions of STEM subjects develop through interactions between conceptual knowledge, classroom experiences, and affective evaluation \citep{brown2011understanding, osborne2003attitudes}. Across educational stages, repeated exposure to disciplinary content and institutional routines can stabilise into relatively enduring orientations toward such subjects, shaping beliefs about difficulty and personal competence \citep{mcclure2017stem, olsen2015predicting}. Importantly, these orientations are not purely cognitive: emotions such as anxiety or threat can interfere with learning and engagement. These emotions can crystallise around quantitative coursework and performance expectations \citep{foley2017mathanxiety, luttenberger2018spotlight} and interfere with conceptual understanding and undermine sustained engagement \citep{mallow2006science, stella2022network}. In fact, previous research has shown that affective processes are a core factor in the development of STEM perceptions \citep{osborne2003attitudes, pekrungoetzperry2002}.

A complementary view treats these orientations as \textit{mindsets}: structured, domain-specific systems of meaning and affect that guide how learners interpret challenges and anticipate outcomes \citep{stella2019forma, yeager2012mindsets}. In this work, we use the term \textit{mindset} to denote the organised system of meanings and affective orientations through which individuals interpret and anticipate experiences in a given domain \citep{stella2019forma}. Because mindsets are partly implicit, self-report questionnaires may overlook how emotions and concepts co-occur within the mental lexicon. Cognitive network science offers tools to represent such organisation explicitly by modeling concepts as nodes connected by empirical associative links \citep{stella2021mapping}.

\subsection{Cognitive network science and behavioural forma mentis networks}
Network approaches have been successfully applied to study creativity \citep{beaty2023associative}, curiosity \citep{zurnbassett2018curiosity}, and bias in large language models \citep{abramski2023cognitive}, demonstrating how differences in network organisation reflect differences in cognition, learning, and evaluation \citep{stella2022network}. Cognitive network science models the mental lexicon as a network of concepts connected by relations such as free associations \citep{stella2021mapping}. This lexicon is not a static repository, but a dynamic structure that supports processes of lexical access, memory retrieval, and learning \citep{stella2020forma}. Within this framework, a mindset corresponds to a specific semantic-affective configuration of the cognitive network \citep{stella2020forma, stella2021mapping}.

Behavioural forma mentis networks (BFMNs) operationalise this idea, combining free-association structure with affective (e.g., valence) labels, enabling the study of local semantic neighbourhoods (semantic frames) and their emotional composition, and so the analysis of how concepts are framed within the mental lexicon at the group level \citep{semeraroEmoAtlasEmotionalNetwork2025, stella2019forma}. In these networks, nodes represent concepts, edges represent associative links, and each node is annotated by its perceived valence. This representation captures not only which concepts are connected, but also how they are emotionally evaluated. For example, the concept \textit{mathematics} may be linked to "numbers" and "physics" while being embedded in a neighbourhood containing "anxiety" or "challenge", each with its own emotional polarity \citep{abramski2023cognitive, stella2021mapping}. The resulting structure constitutes a measurable representation of a group's mindset toward mathematics.

Behavioural forma mentis networks are representational models of the mental lexicon rather than neural models \citep{stellaFormaMentisNetworks2020, stella2019forma}. They differ from textual forma mentis networks, which encode syntactic or co-occurrence relations extracted from written corpora \citep{semeraroEmoAtlasEmotionalNetwork2025}. Because BFMNs are grounded in behavioural data from human memory, they are particularly suited to study educational mindsets and affective biases.

A key construct supported by this framework is the valence aura, understood as the local balance of positive, neutral, and negative concepts in the immediate semantic neighbourhood of a target word \citep{stella2019forma, abramski2023cognitive}. Notably, a concept can be judged neutral when considered on its own, yet still sit within a largely negative aura if most of its associates are negatively valenced. This nuance is particularly important in educational settings. Students might not rate “exam” as strongly negative explicitly, even though their related associations tend to concentrate near anxiety, fear, and failure. By modelling both network topology and valence together, BFMNs help to reveal these local affective fields. More recent developments also integrate other lexical features — such as concreteness (see \citet{brysbaert2014concreteness}) and emotion distributions — to further characterise how concepts are mentally represented across different groups \citep{semeraroEmoAtlasEmotionalNetwork2025, stella2022network}.

As previously shown, associative and affective traces reveal interconnected systems of meaning linking STEM subjects to ideas such as complexity, failure, curiosity, or utility, each carrying positive, neutral, or negative emotional tones \citep{stella2019forma, stella2020forma}.
Prior work has also shown that different populations can share an overall positive view of \textit{science} while simultaneously embedding core quantitative subjects (e.g., mathematics, physics, statistics) in more negative neighbourhoods \citep{stella2019forma}. Such patterns provide a network-level lens on how educational experiences may generate evaluative and motivational inconsistencies, with potential implications for anxiety and disengagement. In parallel, recent studies indicate that large language models (LLMs) can reflect population-level semantic biases, making them useful benchmarks for comparing human and artificial representations when appropriately prompted \citep{abramski2023cognitive, ciringione2025math}.

\subsection{Aims} \label{aims}
Building on prior work, this study takes a cognitive network science approach to examine how STEM mindsets are organised across educational settings and anxiety profiles. We also specifically focus on math anxiety, which may emerge as a context-dependent and temporary reaction to evaluation (state anxiety) or as a more enduring, dispositional orientation toward quantitative domains (trait anxiety) \citep{saviola2020trait, spielberger1970manual}. We propose that these components correspond to distinct network signatures, also according to Spielberger's state-trait anxiety distinctions: context-linked anxiety should appear in concrete, evaluation-centered associations, whereas more stable anxiety should manifest as an abstract, pervasive and negative embedding in the network.

\subsection{Hypotheses and research questions}
We advance three hypotheses. First, we expect an evaluative mismatch in students’ semantic frames, such that science and research could be represented in a positive and more concrete way (as more related to empirical outcomes), while core quantitative STEM subjects may lie in more negative affective neighbourhoods, especially for students reporting higher math anxiety. Second, when compared with appropriate random baselines, we expect high-anxiety frames to show lower concreteness. Third, we anticipate that GPT-oss based simulations will probably capture the overall valence patterns observed in human data, but could under-represent the concreteness ratings, the experiential grounding and context-sensitive affect found in human counterparts.

To test these hypotheses, we analysed BFMNs across several student groups and expert participants, directly comparing subgroups defined by educational background and math-anxiety profiles \citep{franchino2025network}. We focused on target words/concepts related to STEM subjects (e.g. mathematics) and educational contexts (actors, places and research), and contrasted human representations with GPT-oss simulations designed and prompted to mirror the same profiles \citep{abramski2023cognitive}. In line with these aims, we report our findings across three studies, focusing on:

\begin{itemize}
    \item perceptions of STEM subjects
    \item educational actors, research, and institutions
    \item concreteness of STEM-related semantic frames
\end{itemize}

Within this framework, we address three research questions:
\begin{itemize}
    \item RQ1 – How do academic populations frame core STEM subjects across educational stages and math-anxiety profiles, and do they show a STEM-science dissonance?
    \item RQ2 – How do STEM and educational or research contexts (actors, places) differ in their network structure and features (e.g., concreteness ratings and valence aura)?
    \item RQ3 – How do these human mindsets compare with corresponding GPT-oss (LLM) based networks, and what does this comparison reveal about the diffusion of educational mindsets into LLMs?
\end{itemize}

\section{Methods}
\subsection{Participants} \label{participants}
We gathered both human participants through convenience sampling and GPT-oss simulated counterparts. In total, the underlying dataset includes $N = 994$ observations across groups. Human samples include early-career STEM \textit{Experts} (59) and \textit{High Schoolers (South)} (159) from \citet{stella2019forma}, alongside additional student groups and their GPT-oss simulations (Table \ref{tab: participants}). All the human participants have been recruited via convenience sampling (specifically through the department's social media and word of mouth). For this reason, the resulting groups reflect accessibility and availability rather than probabilistic selection. Table \ref{tab: participants} summarises all groups included in the study. They were provided informed consent and were not compensated. The study protocol received approval from the Research Ethics Committee of the University of Trento (Protocol number: 2024-039).

We performed data cleaning across all our samples, and excluded participants with 1/3 or more of the data missing and those falling in the median value according to the distribution of the obtained scores at the MAS-IT questionnaire. This questionnaire was used to separate some groups by anxiety level (low/high; see Section \ref{samples_division}, \textit{\nameref{samples_division}}). The number of participants reported in Table \ref{tab: participants} represents the participants who were kept for the analysis.
\begin{table}[!htbp]
    \centering
    \begin{tabular}{l|c|c|c|c}
    \toprule
    Sample & Total & High anxiety & Low anxiety & Excluded \\
    \midrule
    Experts & 59 &  & & \\
    High Schoolers (South) & 159 &  & & \\
    Physics Undergraduates & 10 & & & \\
    GPT-oss Physics Undergraduates & 10 & & & \\
    High Schoolers (North) & 62 & 30 & 27 & 5 \\
    GPT-oss High Schoolers & 62 & 28 & 29 & 5 \\
    Psychology Undergraduates & 316 & 150 & 151 & 15 \\
    GPT-oss Psychology Undergraduates & 316 & 143 & 157 & 16\\
    \bottomrule
    \end{tabular}
    \caption{Groups and subgroups (by anxiety level) used to create behavioural forma mentis networks.}
    \label{tab: participants}
\end{table}

\subsubsection{Human samples}
The \textit{Experts} sample comprises 59 researchers from varied backgrounds, all with advanced STEM training and active research positions. The \textit{High Schoolers (South)} group includes 159 high school students in their final year; we explicitly label them by geographical location ("South") because these participants were recruited from different schools and assessed with distinct procedures (questionnaires and tasks) than the \textit{High Schoolers (North)} group. For full details on the first two groups, see \citet{stella2019forma}.

As reported in Table \ref{tab: participants}, we also collected student samples from several disciplines along with their GPT-generated counterparts: (i) 10 physics undergraduates enrolled in the Physics Department at the University of Trento; (ii) 62 final-year high school students (including both applied-science students and students that study languages, to capture diverse socioeconomic backgrounds and STEM proficiency); and (iii) 316 psychology undergraduates. Data from psychology undergraduates were gathered across multiple years (2023-2025) using slightly different materials (see Section \ref{materials}, \textit{\nameref{materials}}); nevertheless, all members of this cohort were enrolled in the Psychology Bachelor's program at the University of Trento.

The students' participant samples were composed by adults and native Italian speakers. High school students were aged 17-19, while psychology and physics undergraduates went from 18 to 29 years old (mode = 20); age was not available for every participant. Experts were aged 24-39 (mode = 29), had advanced English proficiency (language also used for the experiment), with a roughly balanced gender distribution (56\% male, 44\% female).

\subsubsection{GPT-simulated samples}
Using an LLM, we created a digital counterpart for each human sample except the \textit{Experts} and \textit{High Schoolers (South)}. Specifically, we used GPT-oss to emulate participants and have the model perform the same tasks. We implemented a Python-based simulation that called the OpenAI API with the GPT-oss model (\texttt{gpt-oss-20b}) to generate synthetic responses simulating human answers.

Through iterative prompt refinement we balanced output consistency and variability to better reflect human behaviour. Each simulation received an assigned profile as context for responses across tasks, improving the ecological validity of the synthetic data. Assigned socio-demographic and educational attributes were informed by literature on math anxiety \citep{stella2022network} and included:
\begin{itemize}
    \item Gender: limited to "male" and "female", following prior findings on gender differences in math anxiety responses \citep{beilock2010female};
    \item Age: randomly sampled within plausible ranges for late high school and university students (18-25), since attitudes toward STEM vary across different age and academic experience;
    \item Education level: distinguished final-year high school students from those in the three years of a BSc (psychology or physics) to account for shifts in STEM perceptions across academic progression;
    \item Socioeconomic conditions: randomly assigned to one of five categories - low ("basse"), medium-low ("medio-basse"), medium ("medie"), medium-high ("medio-alte"), or high ("alte") - to reflect background influences on access and attitudes toward education.
\end{itemize}

The personification prompt template applied was:
\begin{quote}
    \textit{Sei un}\texttt{\{gender\}} \textit{student}\texttt{\{gender\}} \textit{italian}\texttt{\{gender\}} \textit{di} \texttt{\{age\}} \textit{anni. Sei iscritt}\texttt{\{gender\}} \textit{al} \texttt{\{year\}} \textit{anno di} \texttt{\{education\}}\textit{. Sei cresciut}\texttt{\{gender\}} \textit{e vivi in condizioni socio-economiche} \texttt{\{socioeconomic\}}\textit{. Pertanto, ricorda che le risposte da fornire nel compito devono essere originali, creative e coerenti con le tue caratteristiche uniche.}
\end{quote}

An English translation of this prompt is provided in \ref{prompt}, \textit{\nameref{prompt}}.

After generating responses with GPT-oss, we randomly sampled the synthetic data to match the human group sizes for physics undergraduates, High Schoolers (North), and psychology undergraduates (see Table \ref{tab: participants}).

\subsubsection{Samples division} \label{samples_division}
Table \ref{tab: participants} shows that the samples of northern high schoolers, physics and psychology undergraduates, and their GPT-oss counterparts were divided into two subgroups ("Low Anxiety", "High Anxiety") plus an "Excluded" category. This split was based on each participant’s overall score on the Math Anxiety Scale (MAS-IT), described in Section \ref{materials} (\textit{\nameref{materials}}). The cutoff was the median MAS-IT score computed within the relevant sample (for example, across the 316 psychology students). Participants whose MAS-IT score equalled the median were excluded from subgroup analyses; those scoring below the median were assigned to "Low anxiety" and those scoring above to "High anxiety". The expert and southern high school samples were not divided because they did not complete the MAS-IT at the time of testing; the physics undergraduates did complete it, but, given their small $N$, were not split into subgroups.

\subsection{Procedure}\label{procedure}
All participants across the samples (both human and GPT-simulated) were asked to complete two main tasks:
\begin{enumerate}
    \item \textit{Free association task:} for each cue word (described later in Section \ref{materials}, \textit{\nameref{materials}}), participants wrote the first three words that came to mind after reading the stimulus. To preserve spontaneity and support the effectiveness of the method, they were instructed to respond as quickly as possible \citep{DeDeyne2013}. If they were unable to produce one or more associations, they were allowed to leave the corresponding responses blank.
    \item \textit{Valence attribution task:} participants then rated the emotional valence of all words involved in the free association task, including both the cue words and their own generated associations (i.e., they only rated associations they had actually provided and skipped any left blank). Ratings were given on a 5-point Likert scale (1 = "very negative", 5 = "very positive", 3 = "neutral"). This task did not present time limits.
\end{enumerate}

In addition, the \textit{Physics} and \textit{Psychology Undergraduates}, the \textit{High Schoolers (North)}, and their GPT counterparts completed the Math Anxiety Scale-IT (MAS-IT) to measure math anxiety. The MAS-IT is the Italian adaptation of the MAS-UK \citep{Hunt2011}. It has been previously validated for assessing math anxiety among Italian undergraduates \citep{franchino2025network}. This instrument assesses three factors: \begin{enumerate} \item Evaluation MA: anxiety experienced during formal evaluations of mathematical ability (e.g., exams, answering publicly).
\item Everyday/Social MA: anxiety in everyday mathematical situations that involve a social component (e.g., managing money, doing mental calculations). \item Passive Observation MA: anxiety elicited by observing math-related activities without active involvement (e.g., watching someone solve problems) \citep{Hunt2011}.
\end{enumerate} Participants indicated their anxiety on a 5-point Likert scale ranging from 1 ("not anxious at all") to 5 ("very anxious").

\subsection{Materials}\label{materials}
The study materials mainly consisted of: (i) the MAS-IT questionnaire described above (for the full set of items translated into Italian, we refer the reader to \citealp{franchino2025network}); (ii) a list of cue words used in the free association task (to gather data for the networks). For complete details on the cue-word sets used in the present studies with the different samples, we refer the reader to Section \ref{cue_words} (\textit{\nameref{cue_words}}).

As reported in Table \ref{tab:cue_words_summary} (\ref{cue_words}), cue words can be organised into conceptually similar categories. In particular, to address our research questions (see Section \ref{aims} \textit{\nameref{aims}}), we selected a limited set of target words (shared across most participant samples) related to topics central to our domains of interest (STEM subjects and educational and research contexts). These are shown in Table \ref{tab:keywords_analysis} and are the target concepts we will then focus on in our networks and analyses.

\begin{table}[!hbpt]
\centering
\begin{tabular}{l|l|l}
\toprule
Study 1: STEM subjects & Study 2: Educational and research contexts (actors and places)\\
\midrule
Mathematics & Professor \\
Statistics & Teacher \\
Physics & Scientist \\
Science & School \\
Computer Science & University \\
Research & \\
\bottomrule
\end{tabular}
\caption{Target words analysed in each experiment, chosen as central to the main domains of interest of the two studies.}
\label{tab:keywords_analysis}
\end{table}

\subsection{Data availability}
All data collected for this study are publicly available on an OSF repository at: \url{https://doi.org/10.17605/OSF.IO/FTGSV}. The repository contains a folder named "database" with four files, comprising the MAS-IT results and the free association and valence-task data, separated by GPT and human samples. In each file, participant IDs (one per row) encode the corresponding sample (e.g., \texttt{gpt\_oss\_psychology\_001} denotes the first GPT-oss-simulated participant from the psychology undergraduates sample). The dataset does not include the high- vs. low-anxiety subgroups; however, these can be reconstructed by computing the MAS-IT median for the entire sample of interest and then classifying each participant in that sample accordingly (as described in Section \ref{samples_division} \textit{\nameref{samples_division}}).

\subsection{Analysis}
We used the Python 3.10.12 programming language for data analysis and for the construction of behavioural forma mentis networks (BFMNs), semantic frames and emotional flowers. Due to variability in group sizes, both parametric and non-parametric analyses were conducted when necessary. For visualisation purposes, some plots in this manuscript featured Italian words as translated into English.
The analyses we are going to present, following previous studies \citep{stella2019forma, stellaFormaMentisNetworks2020, stella2021mapping, ciringione2025math}, aim to examine whether the same concept is potentially perceived differently by the several populations (subgroups) considered here, all coming from academic environments and paths (see Section \ref{participants}, \textit{\nameref{participants}}).

\subsubsection{Behavioral forma mentis networks}
Two main forma mentis network variants are typically distinguished: (i) \textit{Behavioural} FMNs (BFMNs), which reconstruct knowledge from behavioural data such as free association tasks \citep{stella2019forma}; and (ii) \textit{Textual} FMNs (TFMNs), which derive links from syntactic or co-occurrence patterns in written corpora and enrich them with affective data via AI-based parsing and psychological lexicons \citep{stella2020forma}.

\begin{figure}[!htbp]
    \centering
    \hfill
    \centering
    \includegraphics[width=\linewidth]{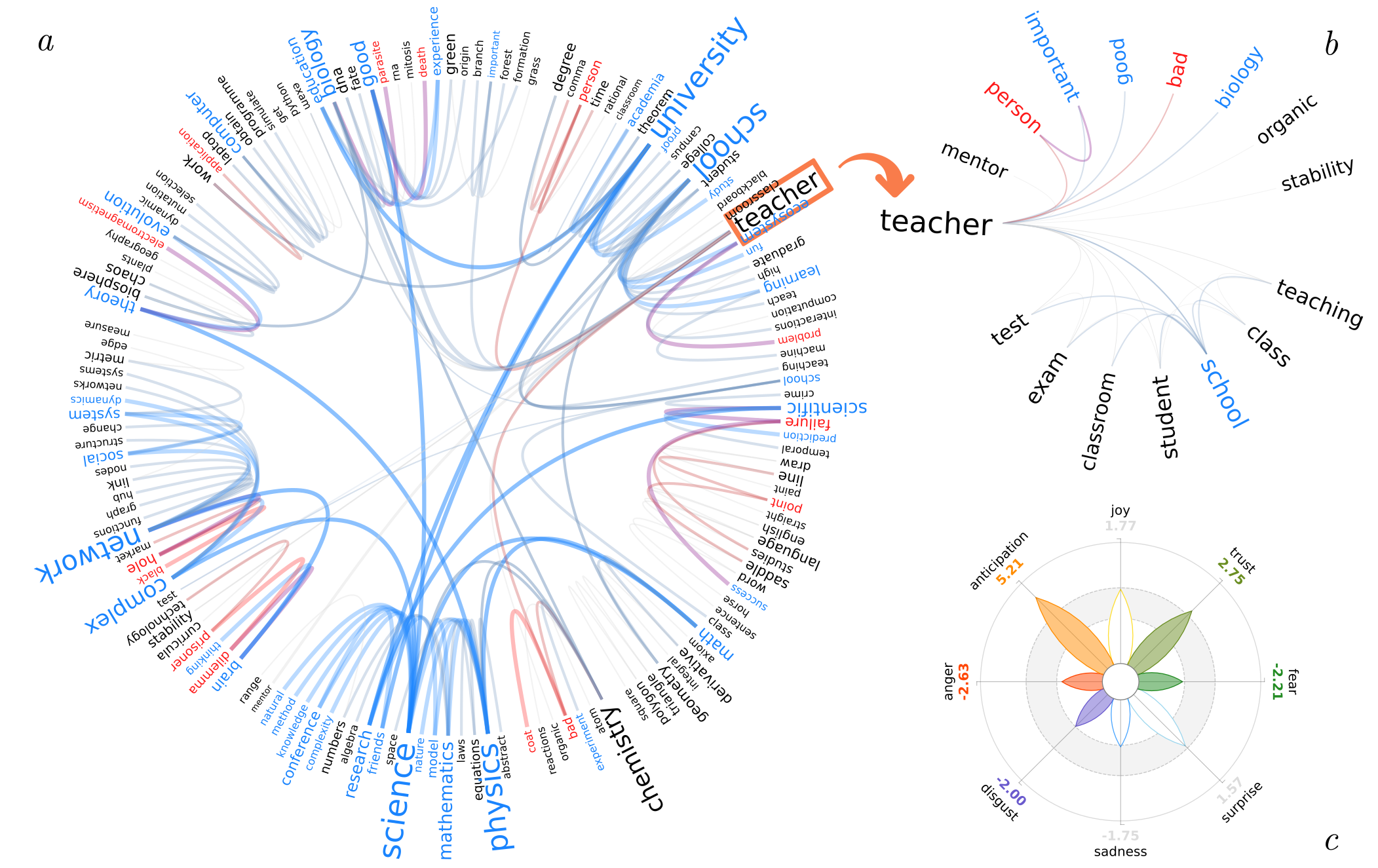}
    \caption{Examples of a behavioural forma mentis network ($a$), a semantic frame (of the node \textbf{\textit{Teacher}}, $b$) extracted from it, and an emotional flower ($c$).}
    \label{fig:example_fmn}
\end{figure}

In this study, we focus on BFMNs (see Figure \ref{fig:example_fmn}) to model mindsets as the joint outcome of cognitive structure (associative organisation) and affective perception (sentiment/valence) \citep{stellaViabilityMultiplexLexical2019, stella2019forma, ciringione2025math}. In other words, BFMNs act as knowledge graphs grounded in psycholinguistic free associations, thereby differing from resources such as WordNet that rely on curated lexical relations rather than behavioural recall links \citep{DeDeyne2013}.

\paragraph{Theoretical Background}
A key strength of BFMNs is their ability to represent knowledge structures via memory recall patterns (i.e., free associations) \citep{DeDeyne2013}, avoiding the need to define specific associative constraints \citep{ciringione2025math}. Free associations can be represented as multidimensional links between memory recall patterns and ideas, combining semantic, syntactic, phonological, and visual, experiential, or simple co-occurrence features of conceptual associations in cognitive memory \citep{abramski2023cognitive}. Because individuals and groups can perceive recalled ideas along a valence continuum from negative to positive, affective perceptions of conceptual valence can alter the structure and quality of free-association networks: for instance, positive concepts may preferentially link with each other, whereas a negative concept might be contrasted with positive ones \citep{ciringione2025math}.

\paragraph{Network data}
Patterns of conceptual associations were collected from individuals completing a free association task, and these associations help reflect how people structure their knowledge around specific concepts \citep{stella2021mapping}. The raw data consisted of cue-response pairs (single words and/or n-grams). Incomplete or empty cue-response associations were discarded prior to network construction. After cleaning, edge lists were created from the association task as lists of cue/association couples, and these were grouped for each subgroup through data pre-processing from row data.

\paragraph{Network construction} \label{network_construction}
Each BFMN created for this study was modeled as a simple, undirected, and unweighted graph ($G=(V,E)$), where $V$ is the set of unique cue/association words (nodes, with valence attributes) and $E$ is the set of unordered pairs $\{u,v\}$ for which at least one valid instance was present in the data. In these networks, nodes represent each cue or associated word with a valence-category attribute (positive, negative, or neutral), and edges correspond to association couples linking concepts. The graphs are undirected because the association between two words has no direction, i.e., $\{u,v\}=\{v,u\}$. Group-level BFMNs were constructed by aggregating the edge lists of all individuals in each subgroup independently, then mapping each node to a specific valence category. A valence category value ($\in\{-1,0,1\}$) was computed from aggregated valence judgments for a specific word within the full dataset (not only within a given participant), using a mean-based criterion and the Kruskal–Wallis test with $\alpha=0.1$ and a minimum number of occurrences per word, as described later in Section \ref{word_valence_analysis} (\textit{\nameref{word_valence_analysis}}). Each numerical valence category value was converted into an interpretable label (\textit{positive}, \textit{negative}, \textit{neutral}) stored as the node \textit{valence} attribute. These networks were treated as unweighted because repeated associations were explicitly collapsed into a single deduplicated simple edge (so each edge between two nodes appeared only once), and no frequency or strength attribute was stored other than \textit{valence} (which was used for word-valence analysis and categorisation, but not for computing network features). Network construction mainly relied on two Python packages, \href{https://networkx.org/}{NetworkX} and EmoAtlas \citep{semeraroEmoAtlasEmotionalNetwork2025}. EmoAtlas is an emotional network analyser of texts that merges psychological lexicons, artificial intelligence, and network science, and provides functionality to build forma mentis networks and related visualisations. We used the NRC Emotional Lexicon (EmoLex), which is the default in EmoAtlas. This procedure yielded 12 networks, one for each subgroup (see Table \ref{tab: participants}, \nameref{samples_division} and \nameref{materials} for details on experimental groups and anxiety-based subdivisions).

\subsubsection{Semantic Frame Analysis}
Semantic frame theory \citep{fillmoreFrameSemanticsText2001} holds that the meaning of a concept can be reconstructed from its associations \citep{semeraroEmoAtlasEmotionalNetwork2025}. Here, a \textit{semantic frame} (Fig.~\ref{fig:example_fmn}) is defined as the neighbourhood of a target node in a forma mentis network, i.e., the set of concepts directly connected to it (cf.~\citep{semeraroEmoAtlasEmotionalNetwork2025}; see also \href{https://github.com/MassimoStel/emoatlas/wiki/2-%E2%80%90-Semantic-Frame-Analysis-with-EmoAtlas}{EmoAtlas documentation}).

After the networks construction, frames were extracted for the target keywords chosen for each study (Section~\ref{materials}, \textit{\nameref{materials}}). Frame extraction and visualisation were performed in Python using EmoAtlas built-in functions \citep{semeraroEmoAtlasEmotionalNetwork2025}. For brevity, we refer to semantic frames simply as “frames”.

We distinguish whole-network analyses from frame-level analyses. Whole-network analyses include (i) word-valence categorisation for all nodes and (ii) any full-network features. All other analyses (e.g., valence aura, Emotional Flowers, and frame visualisations) were conducted at the frame level.

\subsubsection{Statistical Analysis of Word Valence} \label{word_valence_analysis}
Valence scores were obtained from participant ratings of both cue and associated words (Section~\ref{procedure}, \textit{\nameref{procedure}}). To categorise each concept as positive, neutral, or negative, we used a pooled, non-parametric Kruskal-Wallis (KW) test (\href{https://docs.scipy.org/doc/scipy/reference/generated/scipy.stats.kruskal.html}{SciPy documentation}), suitable for independent samples and small group sizes.

Following \citet{stella2019forma}, we implemented a two-tailed KW-based procedure. For each word \(w_i\), we compared the distribution of its valence scores against the pooled distribution of all other valence scores in the same group (i.e., all words except \(w_i\)). If \(p < \alpha = 0.1\) (chosen due to small sample sizes in some groups), \(w_i\) was considered significantly different from the group baseline and labelled \textit{negative} (\textit{positive}) when its mean valence was lower (higher) than the baseline; otherwise it was labelled \textit{neutral}. Because KW requires at least three observations, words occurring fewer than three times, or without ratings, were assigned neutral valence.

Valence ratings were aggregated within each group so that each node received a single group-specific valence label, preserving inter-group differences in valence perception.

\subsubsection{Valence Auras and Affect Analysis}
The \textit{valence aura} (or emotional aura) is a construct introduced in the forma mentis framework \citep{stella2019forma} to capture the emotional context surrounding a concept, beyond its individual rating \citep{stella2019forma, stellaFormaMentisNetworks2020}. Prior work has used auras around academic concepts (e.g., “computation”, “mathematics”, “physics”) to identify potentially anxiety-eliciting content, especially when associations involve methodology, tools, exams, and grades \citep{stella2019forma, stellaFormaMentisNetworks2020, stella2021mapping}.

Aura is defined as the sentiment composition of a word’s immediate neighbours in a network (Fig.~\ref{fig:example_fmn}) \citep{stella2019forma}; equivalently, at the frame level, as the modal sentiment polarity among nodes in the frame \citep{stella2021mapping}. Aura polarity is determined from neighbour valence labels (positive/neutral/negative), typically via a majority (mode) rule: a concept has a negative aura when it is linked to more negative concepts than to positive or neutral ones \citep{stella2019forma, stella2021mapping}. An aura can diverge from a node’s own valence: for instance, a positively valenced concept may be embedded in a negative aura \citep{stella2019forma, stellaFormaMentisNetworks2020}. This enables networked models of group mindsets as more complex tools for investigating nuances of meaning and affect compared to basic numerical questionnaires \citep{stella2021mapping, ciringione2025math}.

Aura analyses also provide evidence for \textit{emotional homophily} in the mental lexicon, i.e., the tendency for concepts with similar valence to cluster in associative structure. Empirically, negative words in negative auras were rated as more negative and elicited higher arousal than negative words in other auras \citep{stella2019forma}. Here, we extend this approach to multiple populations and concepts spanning STEM subjects and educational environments. Confirming such results with additional psycholinguistic data would bolster the idea that sentiment polarities of individual words can influence the structural organisation of semantic memory and the perception of knowledge between such populations \citep{stella2019forma}.

\subsubsection{Semantic-frame Overlap Analysis (Jaccard similarity)} \label{jaccard}
To quantify how strongly two target concepts share the same local semantic neighbourhood, we measured overlap between their semantic frames using the Jaccard similarity coefficient. For a given group, let $F(a)$ and $F(b)$ denote the semantic frames (i.e., sets of first neighbours in the group-level BFMN) of two target nodes $a$ and $b$ (Section~\ref{network_construction}, \textit{\nameref{network_construction}}; Section~\ref{study1}, \textit{\nameref{study1}}). We computed:
\[
J\big(F(a),F(b)\big) = \frac{|F(a)\cap F(b)|}{|F(a)\cup F(b)|},
\]
where $J\in[0,1]$, with $J=0$ indicating no shared neighbours (no overlap) and $J=1$ indicating identical neighbour sets. The Jaccard index provides a normalised measure of overlap that is comparable across networks with different degrees.

\subsubsection{Visualisations}
\paragraph{Edge-bundling layouts for network visualisation}
For visual inspection of both entire networks and semantic frames, EmoAtlas supports hierarchical edge-bundling layouts that position nodes on a circular embedding (Fig.~\ref{fig:example_fmn}). Using established conventions \citep{stella2021mapping, ciringione2025math}, we displayed positive nodes in blue/cyan/black, negative nodes in red, and contrastive links between positive and negative concepts in purple. Edge appearance varies by valence and edge type (colour/thickness/transparency), highlighting clusters and relationships (see the \href{https://github.com/MassimoStel/emoatlas/blob/main/emoatlas/draw_formamentis_bundling.py}{EmoAtlas code/documentation}). Font size is scaled by frame-level closeness centrality, a core feature of EmoAtlas and forma mentis analyses \citep{semeraroEmoAtlasEmotionalNetwork2025}. For human psychology students' samples, to reduce visual clutter given larger networks, we are going to show only association pairs occurring at least twice.

\paragraph{Emotional flowers for emotional profiling} \label{emo_flower}
EmoAtlas also provides \textit{Emotional Flowers}, a flower showing eight primary affective states, following Plutchik's psychoevolutionary theory \citep{plutchikEmotionsLifePerspectives2003}: \textit{joy}, \textit{trust}, \textit{fear}, \textit{surprise}, \textit{sadness}, \textit{disgust}, \textit{anger}, and \textit{anticipation}. Petal length reflects the intensity/frequency of each emotion in the analysed input (words/text), with colours and positions fixed by Plutchik’s model. Longer petals indicate more prominent emotions, while shorter or non coloured petals indicate less significant or absent emotions. The petal's length is scaled to reflect the emotion's score, and its colour is assigned accordingly. A neutral central circle is drawn to represent emotional neutrality.

\subsubsection{Emotional flowers scorings}
Emotion scores are computed by EmoAtlas using the NRC Emotion Lexicon \citep{mohammad2013crowdsourcing} and language-specific models \citep{semeraroEmoAtlasEmotionalNetwork2025}. When statistical significance is evaluated (via $z$-scores), non-significant petals are shown unfilled. Flowers thus allow rapid comparison of emotional profiles across texts, groups, or frames; technical details are provided in \citet{semeraroEmoAtlasEmotionalNetwork2025}.

\subsubsection{Network features} \label{network_features}
For each sample, we computed the same set of descriptors both for the full BFMN and for each extracted semantic frame of the target words. Frame-level and network-level measures were computed with \href{https://networkx.org/documentation/stable/index.html}{NetworkX}.

We report:
\begin{itemize}
    \item $N_v$: number of nodes in a graph;
    \item $N_e$: number of edges in a graph;
    \item $C_i$: clustering coefficient \citep{wattsCollectiveDynamicsSmallworld1998};
    \item $l_G$: average shortest path length \citep{newmanNetworksIntroduction2016};
    \item Hubs: nodes in the top 1\% (full network) or top 5\% (frames) of the degree distribution, with degree $k$ reported in parentheses \citep{hillsBehavioralNetworkScience2024, stellaFormaMentisNetworks2020, newmanNetworksIntroduction2016}.
\end{itemize}

The \textit{clustering coefficient} represents the extent to which the neighbours of a specific node are also neighbours of each other \citep{wattsCollectiveDynamicsSmallworld1998}. In other words, it is a measure of how many adjacent nodes to a specific node in a network are connected. For graphs (Section~\ref{network_construction}, \textit{\nameref{network_construction}}), node-wise clustering was computed as
\[
c_u = \frac{2T(u)}{\deg(u)\big(\deg(u)-1\big)},
\]
where $T(u)$ is the number of triangles through node $u$ and $\deg(u)$ its degree (see \href{https://networkx.org/documentation/stable/reference/algorithms/generated/networkx.algorithms.cluster.clustering.html}{NetworkX documentation}).

The \textit{average shortest path length} ($l_G$) corresponds to the average smallest number of links separating any two nodes \citep{newmanNetworksIntroduction2016}, also known as 'mean geodesic distance' ($\ell$). It was computed as
\[
l_G = \sum_{\substack{s,t \in V \\ s \neq t}} \frac{d(s,t)}{n(n-1)},
\]
where $V$ is the set of nodes in $G$, $d(s, t)$ is the shortest path from $s$ to $t$, and $n$ is the number of nodes in $G$ (see \href{https://networkx.org/documentation/stable/reference/algorithms/generated/networkx.algorithms.shortest_paths.generic.average_shortest_path_length.html}{NetworkX documentation}).

\begin{table}[H]
\footnotesize
\centering
\caption{Sample-wise network features of the entire network and for the target semantic frames.}
\label{tab:network_features_total}
\begin{tabular}{lccccp{9.5cm}}
\toprule
Sample & $N_v$ & $N_e$ & $l_G$ & $d$ & Hubs (top 1\% degree in full network) \\
\midrule
Experts & 1489 & 2861 & 4.51 & 10 & \textit{Science} (65), \textit{Physics} (54), System (53), \textit{School} (51), \textit{Mathematics} (47), \textit{University} (47), \textit{Life} (46), Network (44), \textit{Complex} (38), Art (38), \textit{Chemistry} (34), \textit{Model} (31), Theory (31), \textit{Water} (30), \textit{Biology} (30) \\
H-S South & 4149 & 10207 & 4.06 & 7 & \textit{Physics} (113), \textit{Mathematics} (100), Machine (86), Discussion (82), Time (79), Together (79), \textit{School} (77), Art (67), Renewal (65), \textit{Water} (64), Maximum (63), \textit{Science} (63), System (63), Program (62), Tree (61), Teacher (61), Self organization (61), Rock (60), Behavior (58), Transmission (58), Domination (58), Law (57), \textit{Graph} (56), Society (56), Organization (55), \textit{Person} (55), \textit{Chemistry} (54), Situation (54), Gas (54), Rational (54), Skin (54), Study (53), Action (53), Universe (53), Mathematically (52), Discovery (52), Work (52), \textit{Life} (52), Graduated (51), Populations (51), \textit{Cell} (51), Come on (50), Substance (50), Quality (50), Technology (50) \\
H-Anx H-S & 1207 & 2079 & 3.55 & 5 & \textit{Anxiety} (74), \textit{School} (74), \textit{Mathematics} (72), Life (70), \textit{Discovery} (66), \textit{Research} (64), \textit{Stress} (63), \textit{Physics} (62), \textit{Future} (62), \textit{Art} (62), \textit{Failure} (58), \textit{Teacher} (58), \textit{Grade} (58) \\
L-Anx H-S & 1189 & 1987 & 3.61 & 5 & \textit{Challenge} (68), \textit{Physics} (64), \textit{Mathematics} (62), \textit{Future} (62), \textit{School} (61), \textit{Research} (59), \textit{Experiment} (59), \textit{Data} (58), \textit{Art} (57), Life (57), \textit{Discovery} (56), \textit{Stress} (55), \textit{Science} (55) \\
GPT H-Anx H-S & 520 & 753 & 4.16 & 7 & \textit{Average} (37), \textit{Teacher} (31), Degree (28), Behavior (27), \textit{Development} (27), \textit{Scientist} (27) \\
GPT L-Anx H-S & 540 & 773 & 4.12 & 7 & \textit{Average} (42), \textit{Teacher} (35), Behavior (30), Tests (30), Degree (29), \textit{Development} (28) \\
Physics & 639 & 922 & 4.09 & 6 & \textit{Anxiety} (30), \textit{Physics} (30), Life (29), \textit{Laboratory} (28), \textit{Fear} (28), \textit{Stress} (28), \textit{Failure} (28), \textit{Challenge} (28), \textit{Discovery} (28), \textit{Scientist} (28) \\
GPT Physics & 354 & 426 & 5.42 & 11 & \textit{Mind} (17), Tests (17), \textit{Physics} (16), \textit{Teacher} (16) \\
H-Anx Psy & 3475 & 8618 & 3.40 & 5 & \textit{Anxiety} (252), \textit{Challenge} (209), Behavior (198), \textit{Teacher} (196), Work (190), \textit{Curiosity} (184), \textit{Physics} (182), \textit{Model} (181), \textit{Mathematics} (179), \textit{Personality} (173), \textit{Professor} (170), \textit{School} (163), \textit{Exam} (155), \textit{University} (154), \textit{Creativity} (153), \textit{Numbers} (151), \textit{Mind} (150), \textit{Statistics} (149), \textit{Stress} (149), \textit{Emotion} (148), \textit{Knowledge} (148), \textit{Science} (145), \textit{Task} (141), \textit{Wellbeing} (140), \textit{Future} (140), \textit{Psychology} (139), \textit{Biology} (139), \textit{Boredom} (139), \textit{Equation} (136), \textit{Technology} (128), Degree (127), \textit{Therapy} (123), Report (119), \textit{Discovery} (118), \textit{Neuroscience} (116) \\
L-Anx Psy & 3582 & 8518 & 3.44 & 5 & \textit{Anxiety} (222), \textit{Challenge} (202), \textit{Teacher} (202), \textit{Physics} (201), Behavior (199), \textit{Mathematics} (194), \textit{Curiosity} (194), Work (192), \textit{Personality} (186), \textit{Science} (176), \textit{Model} (172), \textit{Professor} (171), \textit{University} (162), \textit{Stress} (160), \textit{School} (160), \textit{Exam} (156), \textit{Emotion} (155), \textit{Future} (155), \textit{Statistics} (153), \textit{Mind} (152), \textit{Equation} (150), \textit{Numbers} (149), \textit{Boredom} (147), \textit{Wellbeing} (147), \textit{Therapy} (145), \textit{Creativity} (143), \textit{Discovery} (140), \textit{Knowledge} (136), \textit{Task} (132), \textit{Psychology} (132), Degree (127), \textit{Biology} (126), \textit{Grade} (124), \textit{Neuroscience} (120), Report (117), \textit{Technology} (117) \\
GPT H-Anx Psy & 1025 & 1685 & 3.66 & 6 & \textit{Average} (104), \textit{Development} (71), \textit{Equation} (68), Tests (67), Try it (65), \textit{Neuroscience} (65), \textit{Teacher} (64), \textit{Physics} (61), \textit{Psychology} (56), \textit{Scientist} (56), Behavior (54), Work (54) \\
GPT L-Anx Psy & 1305 & 1992 & 3.73 & 6 & \textit{Average} (129), \textit{Equation} (77), \textit{Development} (76), Tests (76), \textit{Neuroscience} (74), \textit{Scientist} (70), \textit{Teacher} (68), \textit{Physics} (66), Try it (65), Behavior (61), \textit{Mind} (57), \textit{Psychology} (56), \textit{Fun} (55), Work (55), \textit{Psychotherapy} (55) \\
\bottomrule
\end{tabular}
\end{table}

\subsubsection{Concreteness Analysis} \label{concreteness}
To characterise how different groups identify key STEM and educational concepts, we analysed the concreteness of nodes in the semantic frames extracted from BFMNs. Concreteness indexes the extent to which a concept is perceptible/experiential (concrete) rather than primarily linguistic/symbolic (abstract) \citep{brysbaert2014concreteness}. Concreteness plays a central role in cognitive theories such as Paivio’s dual-coding framework, which posits that concrete words benefit from both verbal and perceptual representations, making them easier to process and recall \citep{paivioImageryVerbalProcesses1971, paivioDualCodingTheory2013}. It has also been used as a node-level attribute to characterise lexical and conceptual organisation \citep{stellaMultiplexModelMental2018}.

\paragraph{Norms, matching, and lemmatisation}
We relied on an Italian-translated version of the English concreteness norms by \citet{brysbaert2014concreteness} (37,058 words and 2,896 two-word expressions), as a reference for our Italian data. Both the translated concreteness dataset and the semantic frame nodes were lemmatised using a model (\textit{it\_core\_news\_lg-3.7.0}) from \href{https://spacy.io/}{spaCy}, so that different inflected forms were mapped onto a single lemma. Valence information was propagated from surface forms to lemmas using a majority-vote rule across all occurrences of the lemma; in case of ties, valence was set to neutral. This procedure ensured that each lemma-level node was associated with a single concreteness score and a consistent, group-specific valence label. All nodes were included in the analysis, including multi-word expressions and idiomatic n-grams (e.g., \textit{computer science}), which were treated as indivisible conceptual units.

\paragraph{Operationalisation and null comparisons} \label{z_test}
For each group-keyword pair, we computed the empirical mean concreteness of the corresponding frame (excluding the target node), $\bar{x}$. We then used a \textit{z}-test to compare $\bar{x}$ to the mean ($\mu_0$) of a null distribution, used as random baseline: for each frame of size $k$, we sampled 300 random word lists of length $k$ from the translated \citet{brysbaert2014concreteness} dataset to obtain such distribution. We quantified deviation as
\[
z = \frac{\bar{x}-\mu_0}{\delta_0},
\]
where $\delta_0$ denotes the standard deviation of the null distribution, $z>0$ indicates frames more concrete than chance and $z<0$ indicating more abstract ones. Given small subgroup sizes in parts of the dataset, we evaluated significance at $\alpha=0.1$ ($|z|>1.6449$). Mathematical notation related to the concreteness tables (Tables \ref{tab:concreteness_more}, \ref{tab:concreteness_less}) is detailed in \ref{appendix:concreteness_notation}.

\section{Results}
Here we present the results of the analyses described above, organised into three studies addressing complementary aspects of academic and STEM-related representations; STEM subjects (1), perceptions of educational actors, educational places and research (2) and an overview of concreteness analyses (3).
\subsection{Study 1: Reconstructing perceptions of STEM subjects with semantic frames} \label{study1}
In this first experiment, we investigate how different samples frame key STEM subjects: \textit{mathematics}, \textit{statistics}, \textit{physics}, \textit{computer science} and the more general concept of \textit{science}. Building on prior evidence that the latter is often framed positively while other STEM subjects are framed negatively \citep{stella2019forma}, we test whether these contrasts generalise across educational paths, anxiety profiles, and GPT-oss simulations.

\subsubsection{Mathematics} \label{math}
Mathematics is a shared educational foundation across STEM subjects but is often framed negatively by non-STEM student samples \citep{stella2019forma, osborne2003attitudes}.

\begin{figure}[!hbpt]
    \centering
    \includegraphics[width=0.8\linewidth]{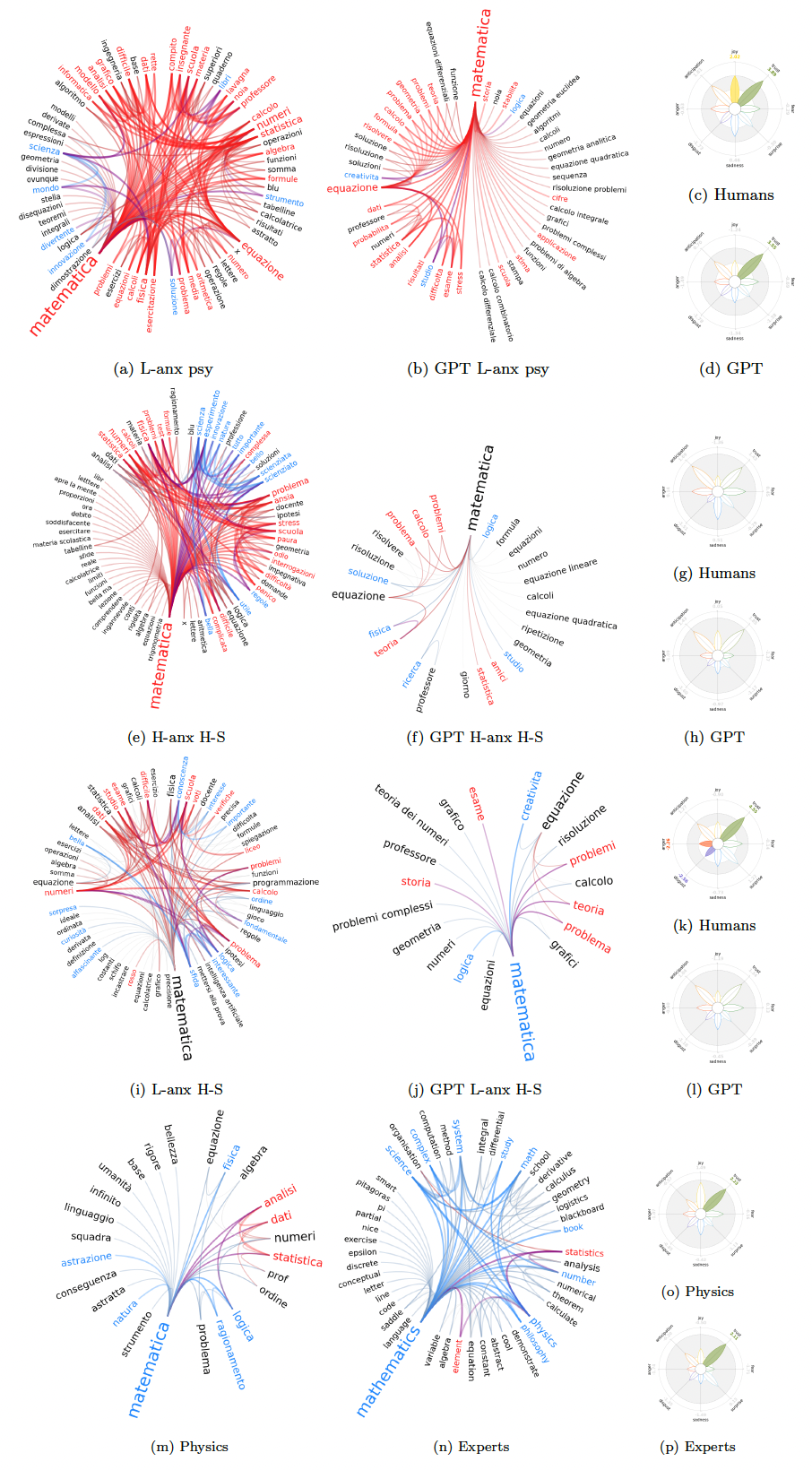}
    \caption{Semantic frames and emotional flowers of \textbf{\textit{Mathematics}}.}
    \label{fig:math_semantic_frames}
\end{figure}

\paragraph{Semantic frame analysis in non-STEM samples} Figure \ref{fig:math_semantic_frames} shows that \textit{mathematics} auras differ sharply across groups (additional frames in \ref{appendix: semantic_frames}, Figure \ref{fig: appen_math_semantic_frames}). Psychology undergraduates (human and GPT) display a strong negative aura, amplified in the high-anxiety subgroup. High schoolers show a clearer low- vs high-anxiety gradient: negativity decreases from high to low anxiety in both human and GPT data. In human high-anxiety subgroups, negativity is specifically tied to emotion- or evaluation-related or terms (e.g., "anxiety", "panic", "hate", "fear", "stress", "exam"), some of which are also network hubs (Table \ref{tab:network_features_mathematics} in \ref{appendix: network_features}).

\paragraph{Semantic frame analysis in STEM samples and LLMs} Physics undergraduates and experts instead frame \textit{mathematics} positively, surrounded by positive/neutral concepts. The few negative associations in physics (e.g., "data", "statistics", "analysis") anticipate discomfort with \textit{statistics}, later confirmed below. Flowers indicate significant trust/joy (L-anx Psy: $z = 3.89$, GPT L-anx Psy: $z = 3.55$, L-anx H-S: $z = 4.55$, Physics: $z = 2.23$, Experts: $z = 2.11$; Figure \ref{fig:math_semantic_frames}), suggesting an \textit{emotional bias} (Section \nameref{emo_flower}) since the associated words would normally evoke trust (according to external lexicon), but participants frame them negatively in context.

\paragraph{Concreteness analysis for mathematics} High-anxiety high schoolers and psychology undergraduates show significantly reduced concreteness relative to null (H-anx H-S: $\bar{x} - \hat{\mu}_0 = -0.26$, $Z = -1.93$, Cohen's $|d| = 0.27$, Cliff's $|\delta| = 0.11$; Psy: $\bar{x} - \hat{\mu}_0 = -0.21$, $Z = -2.26$, Cohen's $|d| = 0.20$, Cliff's $|\delta| = 0.10$; Table \ref{tab:concreteness_less}), indicating that negative, anxiety-linked framings of \textit{mathematics} mirror a relatively abstract perception of it. Experts and physics students do not show this abstractness bias, suggesting a tighter coupling between positive attitudes and concrete representations.

\begin{figure}[!hbpt]
    \centering
    \includegraphics[width=0.45\textwidth]{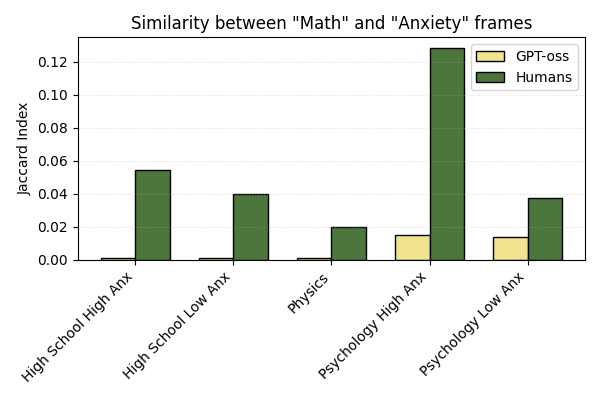}
    \caption{Jaccard similarity values between \textbf{\textit{Mathematics}} and \textbf{\textit{Anxiety}} semantic frames across samples.}
    \label{fig:math_anxiety_comparison}
\end{figure}

\paragraph{Overlap with anxiety} Human networks show substantially higher mathematics-anxiety frames overlap than GPT-oss (see Figure \ref{fig:math_anxiety_comparison} and Section \ref{jaccard} \nameref{jaccard}; Values equal to zero were plotted at $J = 0.001$ for visualisation purposes, so that zero-overlap cases appear as a visible bar). In low-anxiety humans, \textit{math}-\textit{anxiety} similarity is low ($0.02 < J < 0.04$) but rises to $J=0.13$ in high-anxiety psychology undergraduates, up to seven to nine times higher than GPT-oss simulations (<0.02). Thus, anxiety is more integrated into human semantic neighbourhoods of mathematics than in GPT-oss.

\FloatBarrier

\subsubsection{Statistics} \label{statistics}
\textit{Statistics} is a foundational subject for several empirical STEM subjects, including psychology \citep{siew2019anxiety}. This concept appears as negatively framed along \textit{mathematics} across several subsamples (Figure \ref{fig:statistics_semantic_frames}. We provide additional frames and results also in \ref{appendix: semantic_frames}, Figure \ref{fig: appen_statistics_semantic_frames}).

\paragraph{Semantic frame analysis in non-STEM samples} \textit{Statistics} is broadly negative in most groups; psychology undergraduates show the strongest negativity, consistent with a previous study \citep{siew2019anxiety}. Negative hubs and associations include technical terms (e.g., "number", "average", "graphic") and emotion-related terms (e.g., "anxiety", "fear") (Table \ref{tab:network_features_statistics} in \ref{appendix: network_features}). Positive associations (e.g., "university", "psychology", "research", "curiosity") are fewer and highlight perceived outcomes. Low-anxiety high schoolers are the only group framing \textit{statistics} neutrally; high-anxiety high schoolers show stronger negativity, plausibly influenced by \textit{math} as a hub (Table \ref{tab:network_features_statistics} in \ref{appendix: network_features}).

\paragraph{Semantic frame analysis in STEM samples and LLMs} Despite the positive \textit{mathematics} frames of physics students and experts, \textit{statistics} is negatively framed in both samples. Physics students associate it with anxiety-linked terms (e.g., "exam", "panic", "cry"). Experts perceive it negatively but show only one negative association ("regression"). Overall, the pattern supports statistics anxiety \citep{siew2019anxiety} and suggests its persistence across academic paths, with potential pedagogical implications. GPT-simulated students also show a consistent negativity bias towards \textit{statistics}, with even more negative framing both for the target and its associations.

\paragraph{Concreteness analysis for statistics} High-anxiety psychology students and GPT-oss psychology simulations show frames significantly less concrete than random (Table \ref{tab:concreteness_less}), reflecting formal/theory-driven associations (e.g., "probability", "distribution", "analysis") rather than tangible examples.

\paragraph{Emotional bias} As for \textit{mathematics}, \textit{statistics} framings show significant anticipation (L-anx H-S: $z = 2.85$, GPT L-anx H-S: $z = 2.00$; Figure \ref{fig:statistics_semantic_frames}; anticipation can be a component of anxiety according to Plutchik) and sadness (L-anx Psy: $z = 2.08$), in accordance with their negative aura.

\begin{figure}[!hbpt]
    \centering
    \includegraphics[width=0.8\linewidth]{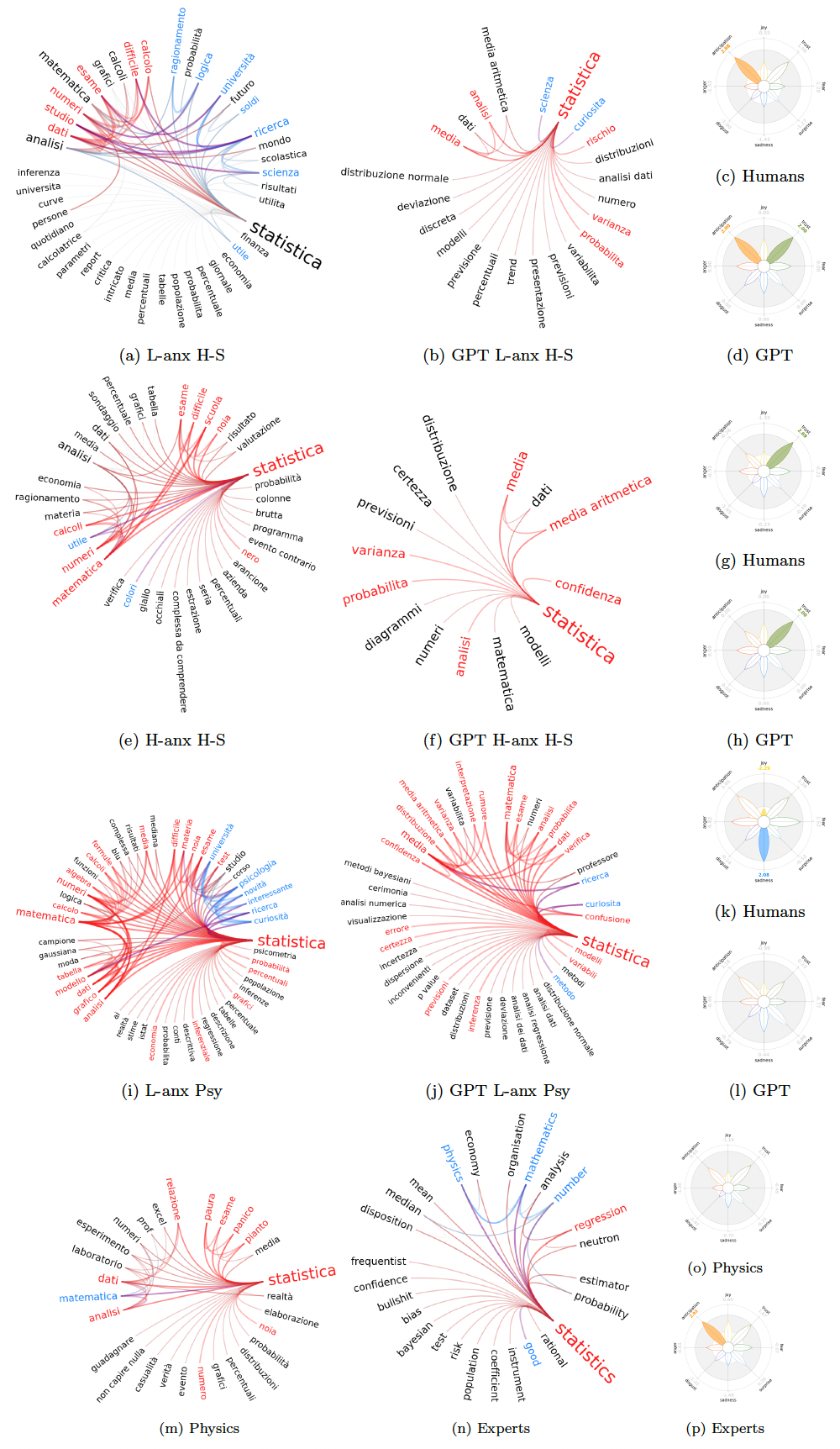}
    \caption{Semantic frames and emotional flowers of \textbf{\textit{Statistics}}.}
    \label{fig:statistics_semantic_frames}
\end{figure}

\FloatBarrier

\subsubsection{Computer science} \label{computer_science}

\textit{Computer science} is a core discipline supporting algorithmic thinking, logic and computational tools for unravelling STEM disciplines \citep{stella2021mapping}. Figure \ref{fig:computer_science_semantic_frames} reports \textit{computer science} frames (others in \ref{appendix: semantic_frames}, Figure \ref{fig: appen_informatics_semantic_frames}).

\paragraph{Semantic frame analysis in non-STEM samples} Psychology undergraduates frame \textit{computer science} negatively (e.g., "difficult", "complicated" and links to "numbers"/"math"). High schoolers are more neutral with some positive associations related to innovation and progress. Physics undergraduates are mostly neutral with a few negative terms (e.g., "data", "fear"), likely partly reflecting fewer associations data.

\paragraph{Semantic frame analysis in LLMs} GPT-simulated students are slightly more positive overall, while mirroring the broad human tendencies in high-anxiety psychology and physics samples. GPT networks contain hubs like "creativity" and "innovation" absent from human hubs (Table \ref{tab:network_features_computer science} in \ref{appendix: network_features}).

\paragraph{Elicited emotions by computer science} Anticipation, which may seggest an experienced sense of anxiety, is significant in several samples (GPT L-anx H-S: $z = 2.00$, H-anx Psy: $z = 2.69$, L-anx Psy: $z = 2.67$, GPT L-anx Psy: $z = 2.88$, Physics: $z = 2.00$).

\paragraph{Concreteness analysis for computer science} Several frames are more concrete than random (Table \ref{tab:concreteness_more}): H-S South ($\bar{x} - \hat{\mu}_0 = 0.47$, $Z = 1.81$, Cohen's $|d| = 0.48$, Cliff's $|\delta| = 0.31$), Psy L-anx ($\bar{x} - \hat{\mu}_0 = 0.38$, $Z = 2.48$, Cohen's $|d| = 0.38$, Cliff's $|\delta| = 0.25$), GPT H-anx H-S ($\bar{x} - \hat{\mu}_0 = 0.61$, $Z = 2.20$, Cohen's $|d| = 0.59$, Cliff's $|\delta| = 0.38$), suggesting \textit{computer science} remains anchored to tangible activities/artefacts even when attitudes are mixed, distinguishing its framing from the more abstract and evaluative profiles found for the other STEM subjects.

\begin{figure}[!hbpt]
    \centering
    \includegraphics[width=0.8\linewidth]{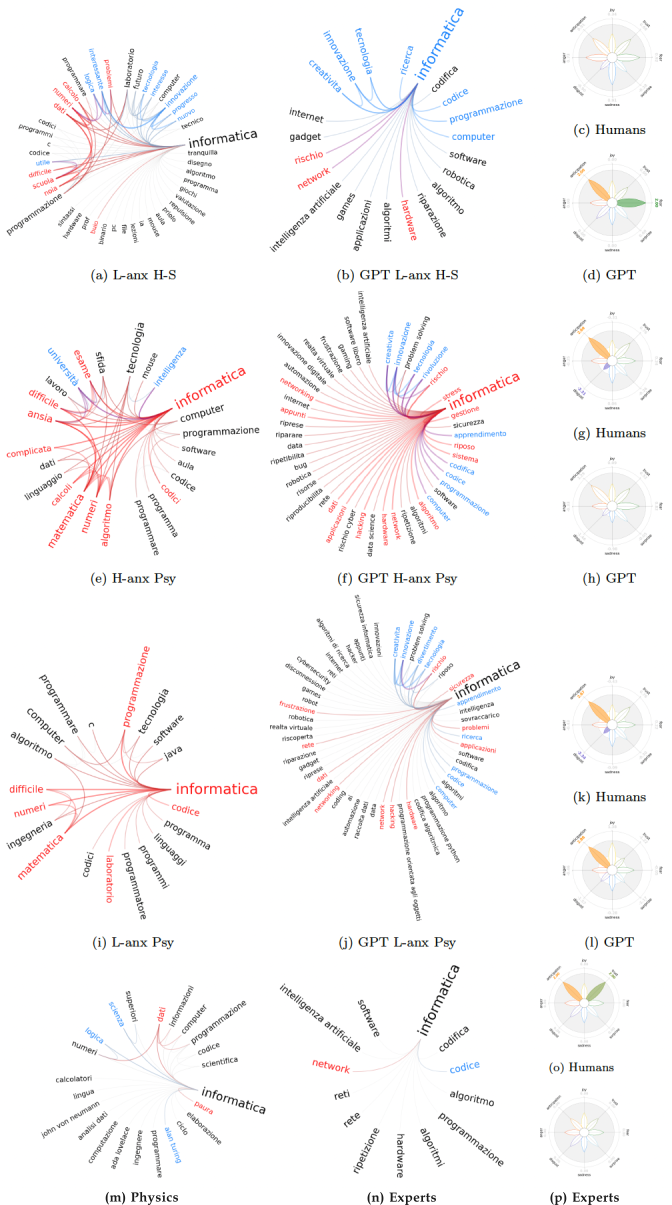}
    \caption{Semantic frames and emotional flowers of \textbf{\textit{Computer science}}.}
    \label{fig:computer_science_semantic_frames}
\end{figure}

\subsubsection{Physics}
Together with maths, \textit{physics} is a core STEM subject, grounded in rigorous scientific methods and theoretical/empirical formalisms \citep{stella2019forma}. Figure \ref{fig:physics_semantic_frames} compares \textit{physics} frames (others in \ref{appendix: semantic_frames}, Figure \ref{fig: appen_physics_semantic_frames}).

\paragraph{Semantic frame analysis in STEM and non-STEM human samples} Experts and physics undergraduates frame \textit{physics} positively, whereas psychology undergraduates and high schoolers show negative or ambivalent perceptions. Low-anxiety high schoolers show neutral frames, while high-anxiety ones show negative frames, a pattern that we also found when exploring the semantics frames of \textit{computer science}. In psychology undergraduates and high schooler frames, negative framings and associations (e.g., "math", "stress", "anxiety") coexists with a positive idealized vision due to the empirical outcome of the subject (e.g., "curiosity", "science", "discovery", "usefulness"), similar to what we'll see for science.

\paragraph{Semantic frame analysis in LLMs} GPT-oss simulations broadly reproduce valence tendencies for psychology undergraduates and low-anxiety high schoolers, but diverge for high-anxiety high schoolers (positive frame) and GPT physics (neutral frame). These mismatches, confirmed by the flowers (human ones tend more towards anticipation, while GPT ones tend towards trust and surprise), further suggest that GPT-oss cannot fully replicate the emotional components of the human frames (tending instead to be more theoretical), the interesting aspect of empirical outcomes, as well as the nuanced influence of the personal experience and pedagogical context (e.g. "problems", "high school", "effort"), which all seem to modulate how anxiety shapes STEM framing.

\begin{figure}[!hbpt]
    \centering
    \includegraphics[width=0.8\linewidth]{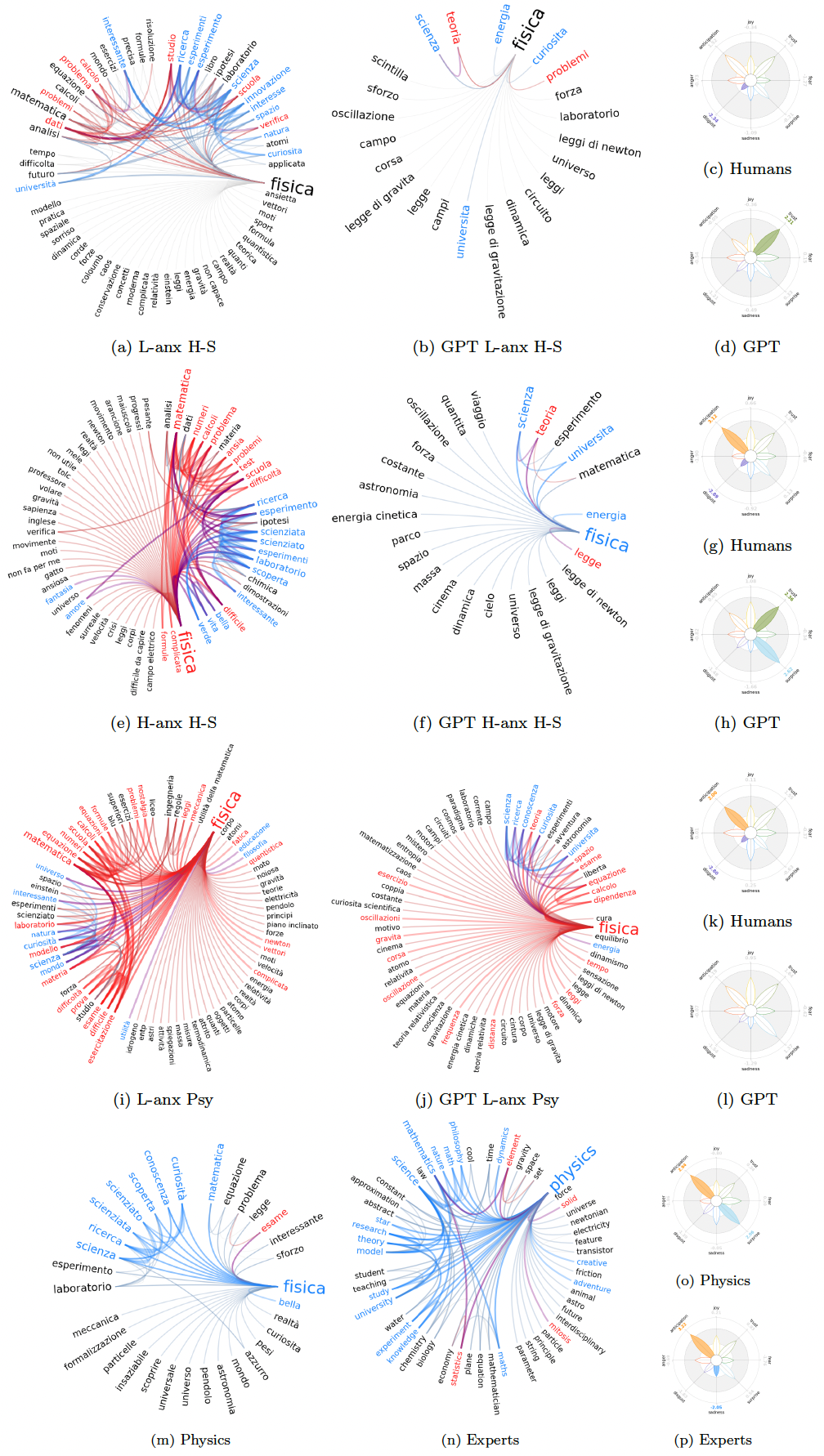}
    \caption{Semantic frames and emotional flowers of \textbf{\textit{Physics}}.}
    \label{fig:physics_semantic_frames}
\end{figure}

\FloatBarrier

\subsubsection{Science} \label{science}
STEM subjects share a core vision of \textit{science} \citep{osborne2003attitudes}. Figure \ref{fig:science_semantic_frames} reports \textit{science} frames (supplementary frames in \ref{appendix: semantic_frames}, Figure \ref{fig: appen_science_semantic_frames}).

\paragraph{Positivity bias for the frame of science} Across groups, \textit{science} is consistently framed positively, contrasting with negativity around other STEM domains. This replicates and extends prior findings that students surround \textit{science} with positive evaluation while assigning negative/ambivalent meaning to its quantitative foundations \citep{stella2019forma}.

\paragraph{Positive associations in the semantic frame analysis for science} All students (human and GPT) link \textit{science} to useful outcomes and interest as with "important", "interesting", "future", "innovation", "progress", "knowledge", "culture", "curiosity". Experts also frame \textit{science} positively, with positive associations more related to academic/technical contexts (e.g., "PhD", "conference", "research", "network", "complex", "physics").

\paragraph{Negative associations in the semantic frame analysis for science} Negative concepts appear mainly in high-anxiety psychology samples (human/GPT) and in southern high schoolers, often as links to other STEM subjects, resulting in hubs (physics, math, biology, statistics).

\paragraph{Emotional patterns associated with science} Trust is significant in several samples: H-anx H-S ($z = 4.09$) with joy ($z = 2.17$) and surprise ($z = 2.22$); H-anx Psy trust ($z = 4.44$) with anticipation ($z = 2.93$) and lack of anger ($z = -2.09$) and disgust ($z = -2.90$); Experts trust ($z = 2.58$) with anticipation ($z = 5.48$) and lack of anger ($z = -2.57$), disgust ($z = -2.09$) and fear ($z = -2.32$); H-S South trust ($z = 3.30$) with lack of anger ($z = -2.39$). GPT high-anxiety high schoolers and psychology simulations show anticipation and surprise, mirroring humans quite well, but only for these emotions. These results align with the broad positive framing of this subject.

\paragraph{Concreteness analysis for science} GPT-oss frames are systematically more abstract: GPT psychology and physics undergraduates and high/low-anxiety high schoolers show mean differences from $-0.40$ to $-0.48$ with $|Z|\approx 2$ or greater, indicating \textit{science} is predominantly framed through theoretical, methodological, or discursive notions (e.g., "theory", "analysis", "knowledge") rather than through concrete laboratory practices, tools, or observable phenomena. This stands in contrast with many human frames, where \textit{science} is instead found to be surrounded by concrete and positive associations.

Taking all these findings together, a coherent pattern of cognitive dissonance emerges. Southern high schoolers, for example, display a slightly more concrete (than random) frame for \textit{science} (Table \ref{tab:concreteness_more}), semantically surrounded by positive, future/outcome-oriented associations. At the same time, the same and closely related populations, especially psychology undergraduates and high-anxiety high schoolers, produced frames for \textit{mathematics} and \textit{statistics} that are both more negative in their valence auras and, in some cases, significantly less concrete than random (see sections \ref{math} \textit{\nameref{math}}, \ref{statistics} \textit{\nameref{statistics}}). Students thus appear to endorse \textit{science} as a useful, inspiring, socially valuable ideal, while simultaneously construing its quantitative backbone and the associated evaluation practices as distant, evaluative and affectively threatening. Overall, science tends to be perceived as less concrete by GPT-oss than humans do, likely related to the experimental and practical "relationship" with the subject the latter have, while GPT-oss frames it on a more theoretical level, missing to fully replicate the human empirically and emotionally grounded experience of it.

\begin{figure}[!hbpt]
    \centering
    \includegraphics[width=0.8\linewidth]{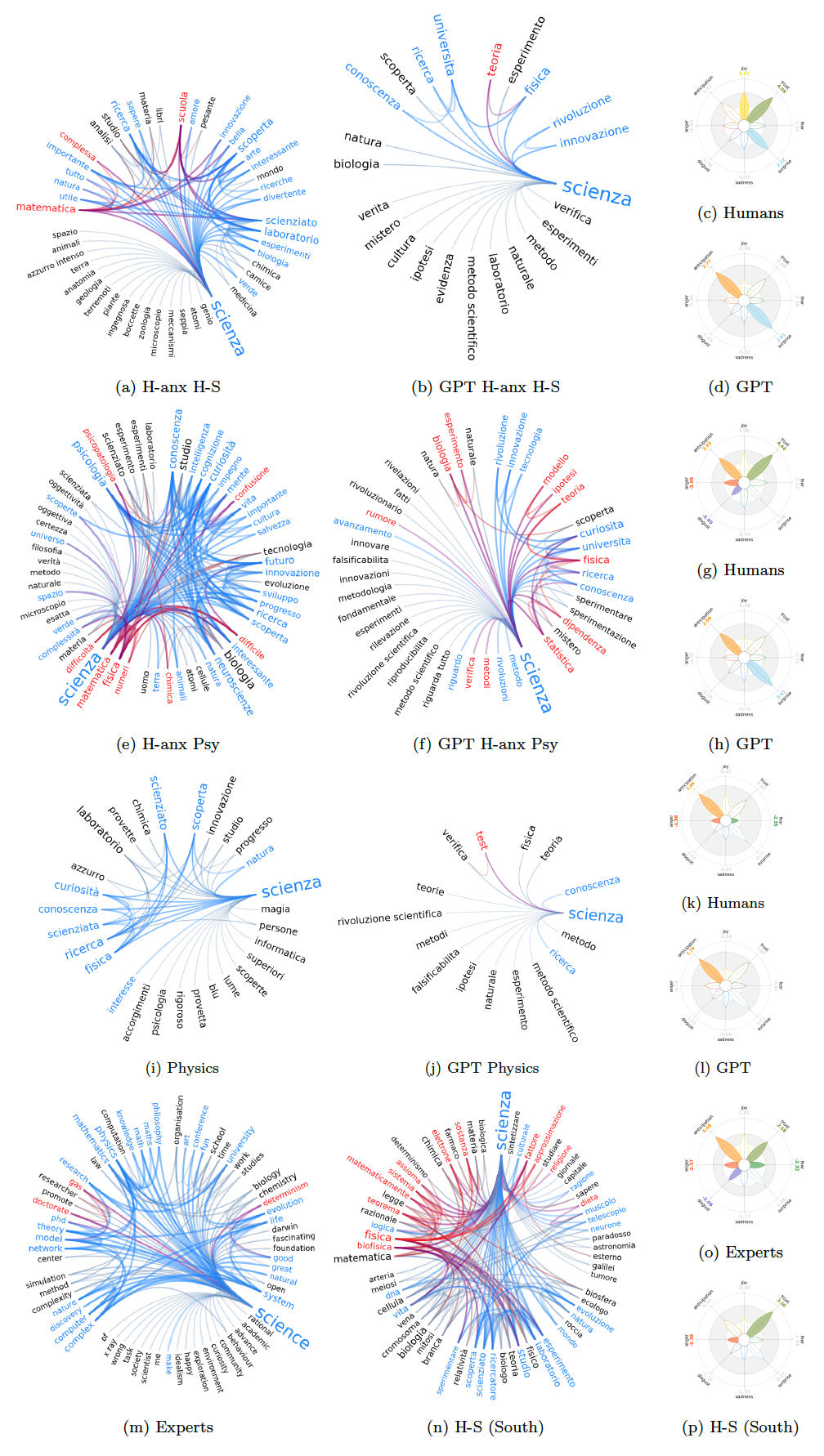}
    \caption{Semantic frames and emotional flowers of \textbf{\textit{Science}}.}
    \label{fig:science_semantic_frames}
\end{figure}

\FloatBarrier

\subsection{Study 2: Perceptions of Educational Actors, Educational Places and Research} \label{study2}
Important aspects of academic life that might influence math anxiety levels are the role of models who guide students in their educational careers \citep{serbati2015implementation} and the educational places \citep{olsen2015predicting} where these take place. Study 2 extends the analysis to educational actors, places, and research, focusing on \textit{professor}, \textit{teacher}, \textit{scientist}, \textit{research}, \textit{school} and \textit{university}. By analysing the semantic frames of these actors, activities, and institutions, this study aims to capture how different samples represent the human, institutional, and research-related dimensions of education, and how these representations may shape the emotional and cognitive climate surrounding academic life. Due to lack of data (associations), some visualizations are not reported for "professor" (experts, physics students, low and high anxiety northern high schoolers) and "teacher" (physics students and low and high anxiety northern high schoolers).

\subsubsection{Academic actors and research}
\begin{figure}[!hbpt]
    \centering
    \includegraphics[width=0.8\linewidth]{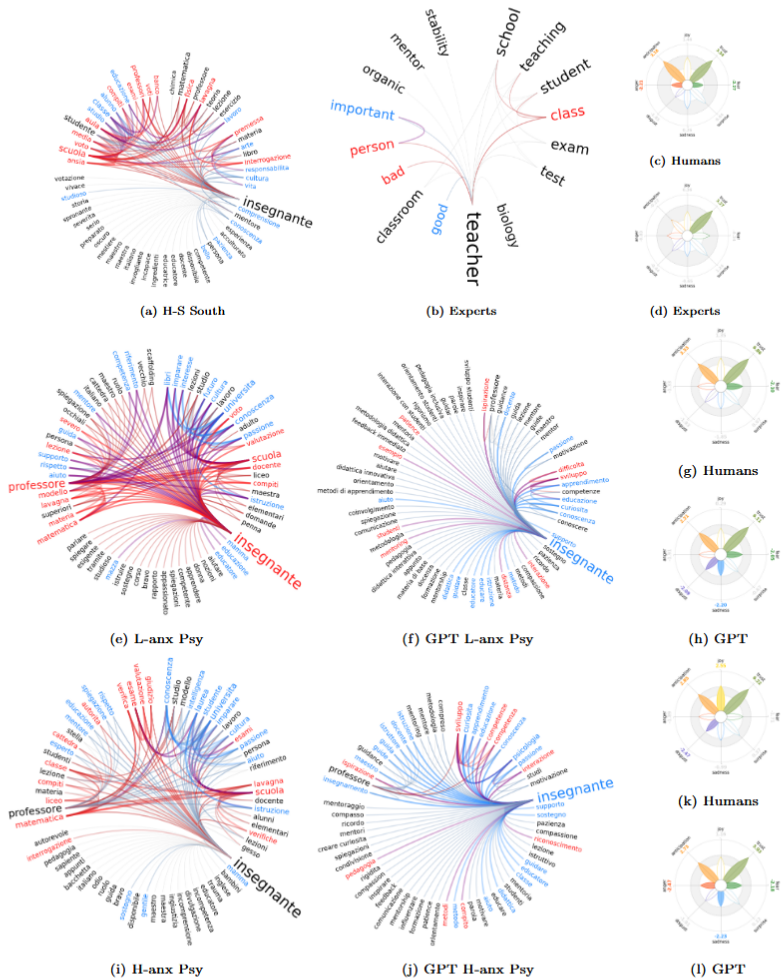}
    \caption{Semantic frames and emotional flowers of \textbf{\textit{Teacher}}.}
    \label{fig:teacher_semantic_frames}
\end{figure}

\begin{figure}[!hbpt]
    \centering
    \includegraphics[width=0.8\linewidth]{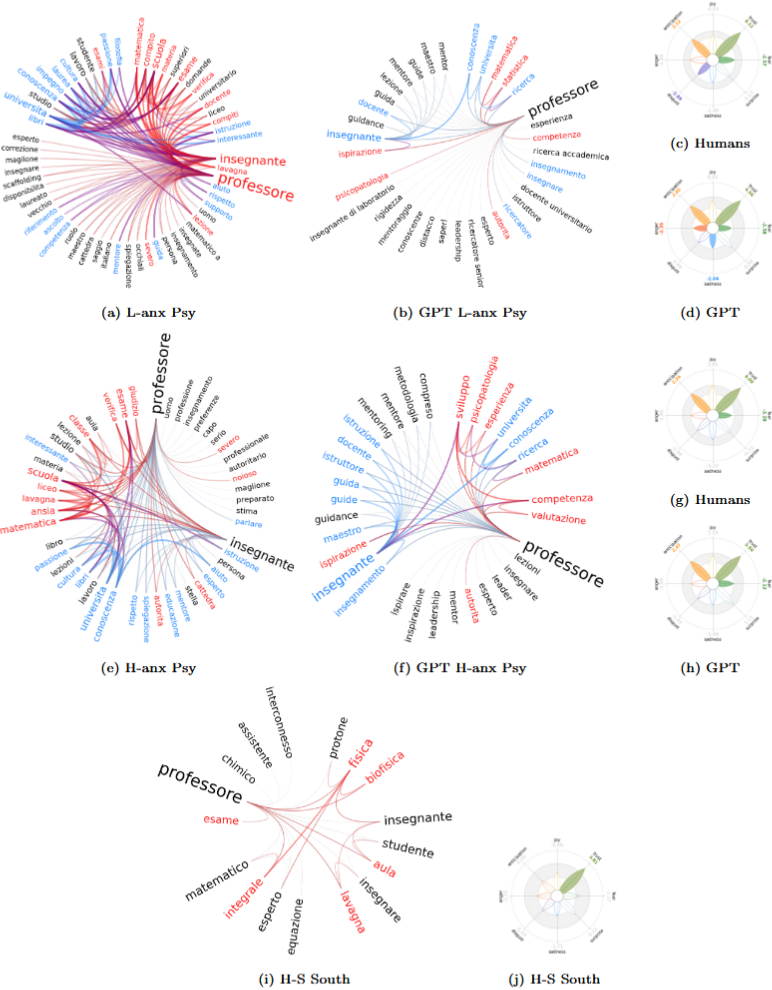}
    \caption{Semantic frames and emotional flowers of \textbf{\textit{Professor}}.}
    \label{fig:professor_semantic_frames}
\end{figure}

\paragraph{Evaluation-driven framing of professor and teacher} In Figures \ref{fig:teacher_semantic_frames} and \ref{fig:professor_semantic_frames}, human psychology students (low/high anxiety) frame both \textit{professor} and \textit{teacher} negatively, while also including neutral/positive associations (e.g., "support", "help", "respect"). Negativity is primarily evaluation-driven and linked to academic environment cues (e.g., "blackboard", "task") and to \textit{mathematics}, which appears among several frames and hubs for both concepts (Tables \ref{tab:network_features_professor} and \ref{tab:network_features_teacher}). GPT-simulated psychology samples do not show this negativity bias (\textit{professor} neutral; \textit{teacher} positive).

\paragraph{Concreteness analysis for professor} The strongest concreteness effect (see Table \ref{tab:concreteness_more}) occurs for \textit{professor} in the southern high schoolers group ($\bar{x} - \hat{\mu}_0 = 0.65$, $Z = 2.62$, Cohen's $|d| = 0.67$, Cliff's $|\delta| = 0.41$), indicating a medium-to-large effect in which the neighbourhood of \textit{professor} is clearly biased toward concrete, experientially grounded concepts. This group anchors the figure of the professor to tangible experiences and settings (e.g., classroom, lessons, physical objects, interpersonal interactions) rather than purely abstract notions of authority, ranking, or institutional roles, and they seem to do it more than the other considered groups.

\begin{figure}[!hbpt]
    \centering
    \includegraphics[width=0.8\linewidth]{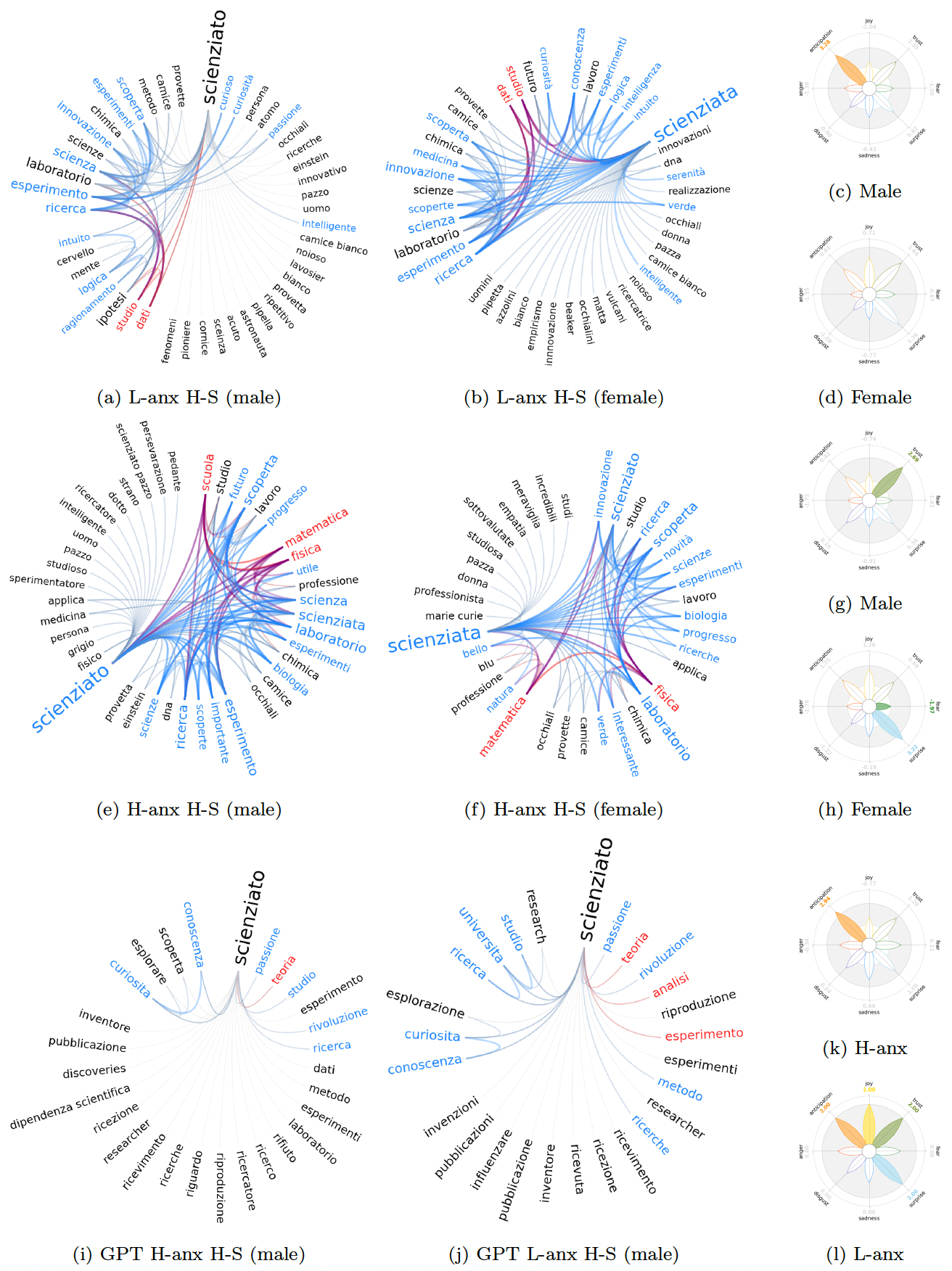}
    \caption{Semantic frames of the node \textbf{\textit{Scientist}}. "Male" and "Female" refer to the Italian form of the word.}
    \label{fig:scientist_semantic_frames}
\end{figure}

\paragraph{From evaluators to researchers: Framing of "scientist"} Relative to \textit{professor}/\textit{teacher}, \textit{scientist} shows a clearer shift toward positive framing (see Figure \ref{fig:scientist_semantic_frames}). High school students generally perceive scientists as constructive, research-focused figures, with negative associations largely tied to academic work and STEM disciplines (e.g., “math”, “physics”, “study”, “data”), while GPT responses tend to be more neutral. Across all human samples, scientists are also stereotypically linked to “crazy”, reflecting obsessive dedication that may convey psychological distance or ambivalence, a bias absent in GPT responses. These patterns suggest that students view scientists less as evaluators and more as actors engaged in meaningful intellectual work.

\begin{figure}[!hbpt]
    \centering
    \includegraphics[width=0.8\linewidth]{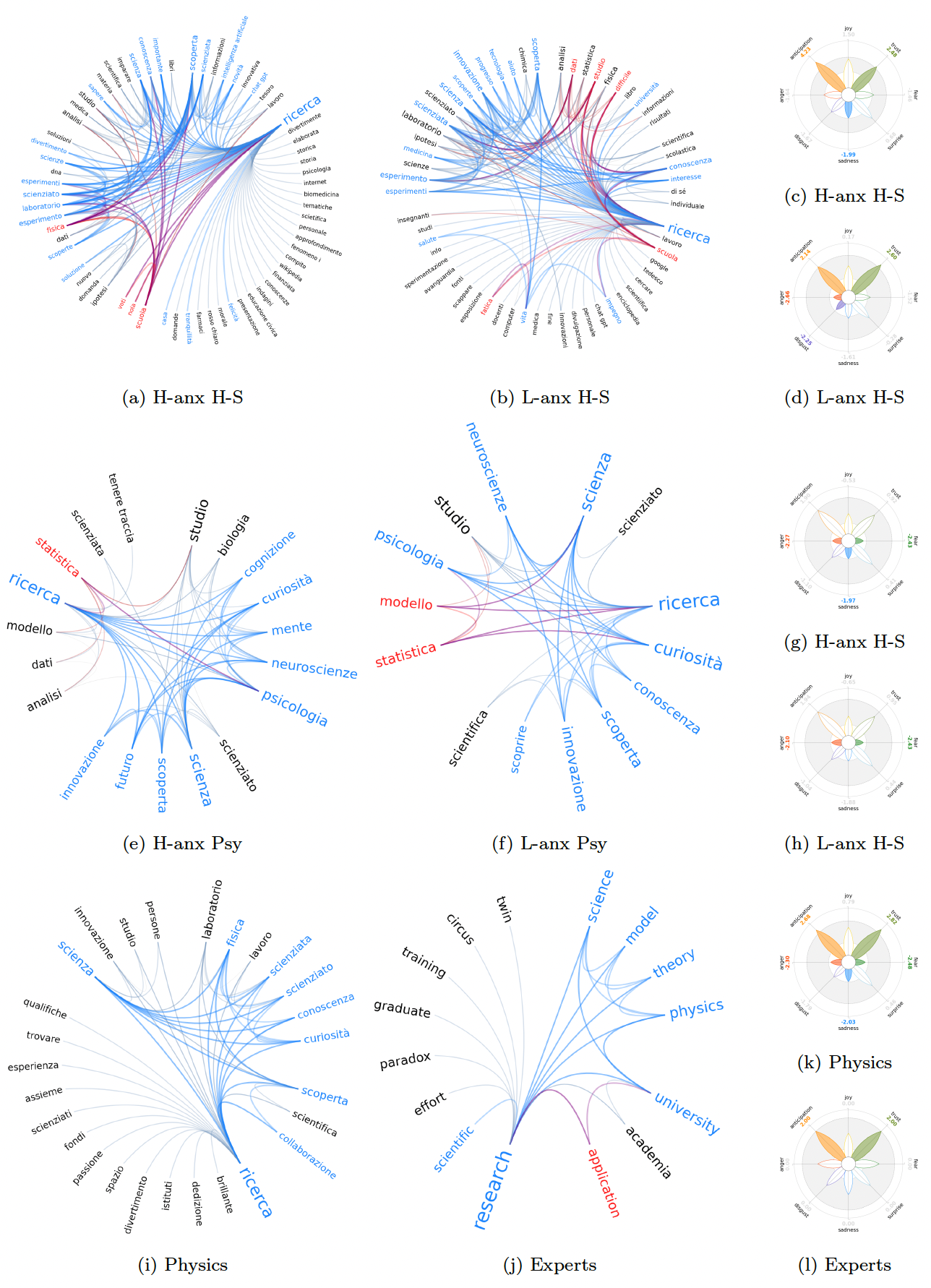}
    \caption{Semantic frames and emotional flowers of \textbf{\textit{Research}}.}
    \label{fig:research_semantic_frames}
\end{figure}

\paragraph{From scientists to research: Forward-looking and concrete representations} \textit{Research} is framed very positively across all groups (Figure \ref{fig:research_semantic_frames}). While some negative associations appear (such as “difficult”, “boredom”, and specific STEM subjects), they are more related to fatigue than anxiety, and research is largely perceived as future-oriented and meaningful. Anxiety triggering associations are largely absent from these frames (also from hubs; Table \ref{tab:network_features_research}). For southern high school students, research also shows high concreteness, suggesting it is imagined through specific activities, tools, and contexts rather than as an abstract ideal. Together, these findings support evidence that students can maintain a positive and concrete image of scientists and research while simultaneously experiencing anxiety toward the quantitative and evaluative aspects of STEM education.

\FloatBarrier

\subsubsection{School vs University} \label{schoolvsuniversity}
\begin{figure}[!hbpt]
    \centering
    \includegraphics[width=0.8\linewidth]{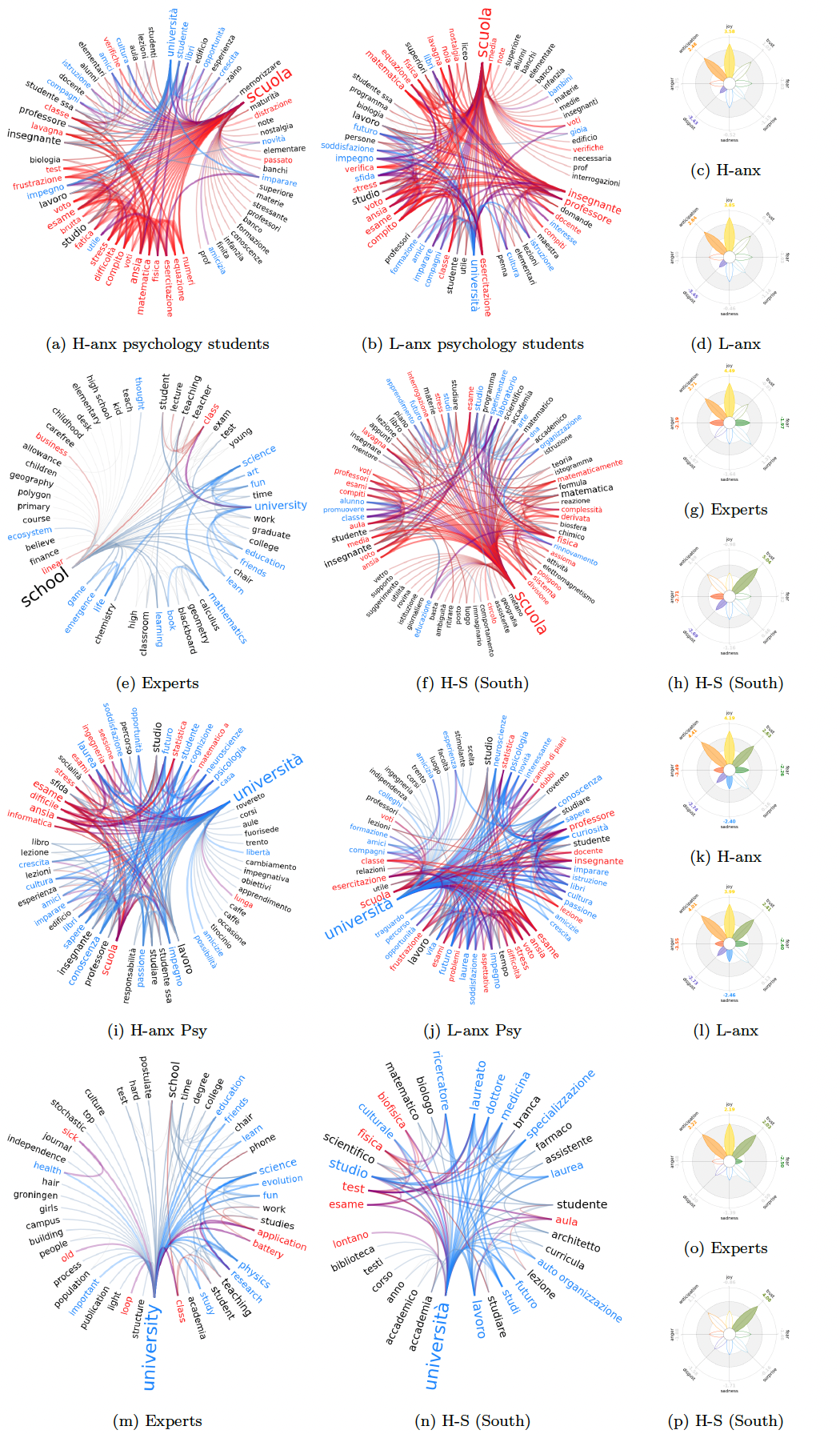}
    \caption{Semantic frames of \textbf{\textit{School}} (top) vs \textbf{\textit{University}} (bottom).}
    \label{fig:school_uni_semantic_frames}
\end{figure}

\paragraph{Emotional contrast in school and university framings} Across most human samples, \textit{university} is predominantly framed positively, whereas \textit{school} tends to elicit negative associations and is framed negatively (Figure \ref{fig:school_uni_semantic_frames}). This pattern is especially pronounced among high school students and psychology undergraduates, who frequently associate \textit{school} with pressure, difficulty, and STEM-related anxiety (for layout reasons, we didn't report the northern high schoolers and gpt frames, but they were very similar to their human counterparts). In contrast, the experts group frames \textit{school} as neutral, associating it mainly with neutral (34) and positive (15) concepts, and only a small number of negative ones (3). This suggests a noticeable shift in perceptions of school across educational stages: those currently enrolled in high school or the early university system evaluate it more negatively than those who are more distant from it.

\paragraph{Shared sources of anxiety and divergent positive associations} For both \textit{school} and \textit{university}, negative associations cluster around evaluation, contextual objects and STEM subjects (exams, grades, teachers, STEM subjects, blackboard...). In \textit{school} frames, "math", "anxiety", "stress" appear among hubs in several samples (Table \ref{tab:network_features_school} in \ref{appendix: network_features}). We notice that, however, anxiety-related terms remain but co-occur with "curiosity", "challenge", "future", and in general positive future-oriented associations, especially in the \textit{university} frames. This means that all samples tend to recognize the good outcomes that attending such environments can bring, with a little contrast between school and university. GPT-simulated students show the same broad contrast.

\paragraph{Concreteness analysis for "school" and "university"} Looking at Table \ref{tab:concreteness_more}, the frame of \textit{school} in southern high schoolers shows a positive mean difference ($\bar{x} - \hat{\mu}_0 = 0.24$, $Z = 2.04$, Cohen's $|d| = 0.26$, Cliff's $|\delta| = 0.18$), and \textit{University} exhibits a similar pattern ($\bar{x} - \hat{\mu}_0 = 0.40$, $Z = 2.18$, Cohen's $|d| = 0.40$, Cliff's $|\delta| = 0.26$). For this sample, both \textit{school} and \textit{university} are grounded in concrete imagery and situations (e.g., classrooms, buildings, schedules, people, specific activities), rather than abstract educational ideals. The \textit{school} frame in the experts group is also more concrete than random ($\bar{x} - \hat{\mu}_0 = 0.38$, $Z = 2.58$, Cohen's $|d| = 0.39$, Cliff's $|\delta| = 0.25$). Such result, together with the emotional nature of the corresponding frame, reinforce the idea that experts conceptualise schooling through vivid, sensory-rich experiences and daily routines, in an emotional state that is positive or neutral, which is in stark contrast with what has been shown by the southern high schoolers. Even if framed as more concrete than random by both groups, the perception of the same concepts (i.e., \textit{school} or \textit{university}) seems to be opposite for the two groups in terms of valence auras. Thus, concreteness can coexist with opposite valence auras across groups.

\FloatBarrier

\subsection{Study 3: An overview of concreteness analyses} \label{study3}
To quantify experiential grounding, we compare frame mean concreteness to null baselines (Section \ref{concreteness}, \nameref{concreteness}), reporting only significant deviations ($\alpha = 0.1$, two-tailed; $|Z|>1.6449$). Tables \ref{tab:concreteness_more} and \ref{tab:concreteness_less} include effect sizes (Cohen's $|d|$, Cliff's $|\delta|$; notation and interpretation details in \ref{appendix:concreteness_notation}). We report absolute effect sizes because direction is conveyed by $Z$; conventional effect size thresholds are followed here \citep{cohenStatisticalPowerAnalysis1988, meisselUsingCliffsDelta2024}, also accounting that typically social-science effects are smaller \citep{ferguson2009effect}.

\paragraph{Frames more concrete than random} Significant positive deviations (Table \ref{tab:concreteness_more}) are concentrated in southern high schoolers (\textit{Professor}, \textit{Computer Science}, \textit{Research}, \textit{University}, \textit{Science}, \textit{School}) and in experts (\textit{School}, \textit{University}), generally indicating experience-grounded representations of institutional actors/places and, for some groups, STEM-related concepts.

\begin{table}[!htbp]
\centering
\small
\begin{tabular}{llrr|rrrr}
\toprule
group & keyword & k & $\bar{x}-\hat{\mu}_0$ & $Z$ & Cohen's $d$ & $\bar{x}$ & Cliff's $\delta$ \\
\midrule
H-S South & Professor & 17 & 0.65 & 2.62 & 0.67 & 3.65 & 0.41 \\
GPT H-S H-anx & Computer Science & 12 & 0.61 & 2.20 & 0.59 & 3.58 & 0.38 \\
H-S South & Computer Science & 15 & 0.47 & 1.81 & 0.48 & 3.46 & 0.31 \\
H-S South & Research & 13 & 0.47 & 1.66 & 0.49 & 3.48 & 0.29 \\
H-S South & University & 30 & 0.40 & 2.18 & 0.40 & 3.38 & 0.26 \\
Experts & School & 51 & 0.38 & 2.58 & 0.39 & 3.36 & 0.25 \\
Psy L-anx & Computer Science & 46 & 0.38 & 2.48 & 0.38 & 3.37 & 0.25 \\
Experts & University & 43 & 0.27 & 1.82 & 0.30 & 3.28 & 0.19 \\
H-S South & Science & 55 & 0.25 & 1.85 & 0.23 & 3.22 & 0.14 \\
H-S South & School & 66 & 0.24 & 2.04 & 0.26 & 3.24 & 0.18 \\
\bottomrule
\end{tabular}
\caption{Frames showing significantly higher concreteness ($|Z| > 1.6449$, $\alpha=0.1$), ordered by decreasing $\bar{x} - \hat{\mu}_0$.}
\label{tab:concreteness_more}
\end{table}

\FloatBarrier

\paragraph{Frames less concrete than random} Significant negative deviations (Table \ref{tab:concreteness_less}) occur mainly in GPT-oss frames and in some human university and high school groups for \textit{Mathematics}, \textit{Statistics}, and \textit{Science}. GPT-oss particularly encodes \textit{Science} and \textit{Statistics} in more abstract, decontextualised terms, and using higher-level, diagnostic, or theoretical terms, reflecting a lexicon dominated by textbook, clinical, or academic discourse rather than experiential content. 

\begin{table}[!htbp]
\centering
\small
\begin{tabular}{llrr|rrrr}
\toprule
group & keyword & k & $\bar{x}-\hat{\mu}_0$ & $Z$ & Cohen's $d$ & $\bar{x}$ & Cliff's $\delta$ \\
\midrule
GPT Phy & Science & 12 & -0.48 & -1.69 & 0.50 & 2.51 & 0.26 \\
GPT H-S H-anx & Science & 20 & -0.46 & -2.05 & 0.44 & 2.56 & 0.24 \\
GPT Psy H-anx & Science & 32 & -0.44 & -2.44 & 0.44 & 2.54 & 0.23 \\
GPT H-S L-anx & Science & 19 & -0.40 & -1.88 & 0.40 & 2.58 & 0.20 \\
GPT Psy H-anx & Statistics & 28 & -0.39 & -2.16 & 0.40 & 2.58 & 0.18 \\
GPT Psy L-anx & Science & 38 & -0.32 & -1.97 & 0.31 & 2.68 & 0.15 \\
GPT Psy L-anx & Statistics & 29 & -0.32 & -1.81 & 0.30 & 2.67 & 0.12 \\
H-S H-anx & Mathematics & 48 & -0.26 & -1.93 & 0.27 & 2.72 & 0.11 \\
Psy H-anx & Statistics & 91 & -0.23 & -2.13 & 0.22 & 2.76 & 0.11 \\
Psy H-anx & Mathematics & 124 & -0.21 & -2.26 & 0.20 & 2.78 & 0.10 \\
Psy H-anx & Science & 107 & -0.18 & -1.91 & 0.18 & 2.81 & 0.09 \\
Psy L-anx & Science & 124 & -0.17 & -1.86 & 0.17 & 2.81 & 0.09 \\
\bottomrule
\end{tabular}
\caption{Frames showing significantly lower concreteness ($|Z| > 1.6449$, $\alpha=0.1$), ordered by decreasing $\bar{x} - \hat{\mu}_0$.}
\label{tab:concreteness_less}
\end{table}

\FloatBarrier

\paragraph{Overview on concreteness results} Overall, these results delineate a distinct pattern in how different academic domains and actors are grounded in memory. Educational settings and figures (e.g., \textit{School}, \textit{University}, \textit{Professor}) frequently emerge as significantly more concrete than random expectation, particularly for high school students and experts, suggesting these concepts are anchored in tangible, daily experiences. In contrast, core quantitative subjects like \textit{Mathematics} and \textit{Statistics} (the most associated concepts to reported anxiety) are framed as significantly more abstract, especially by high-anxiety student groups. Finally, the systematic tendency of GPT-oss to encode concepts such as \textit{Science} and \textit{Statistics} as highly abstract underscores a key divergence: while digital twins may replicate broad valence trends, they currently struggle to reproduce the experiential and sensorimotor grounding that characterises human educational mindsets. For more detailed analyses, see the results sections dedicated to the concreteness analysis of frames in studies 1 and 2.

\section{Discussion}

This study set out to investigate how STEM mindsets are structured across educational stages and anxiety profiles, how these representations extend to academic contexts and actors, and to what extent large language models reproduce such cognitive–affective organisation. The main tools for such investigation are cognitive network science, via behavioural forma mentis networks \citep{stella2019forma}, and digital twins obtained by simulating students/experts via GPT models \citep{ciringione2025math}.

Across nearly one thousand human observations and matched GPT-based digital twins, the results provide converging evidence for a robust STEM–science dissonance (RQ1): while science and research are consistently embedded in strongly positive and often future-oriented semantic frames, their quantitative foundations (particularly mathematics and statistics) are surrounded by more negative, anxiety-laden neighbourhoods, with negativity intensifying and becoming more abstract in high math-anxiety subgroups. Frame-level overlap analyses further show that anxiety is structurally more integrated into human representations of mathematics than in GPT networks, highlighting the experiential embedding of test anxiety, in agreement with past works \citep{ciringione2025math}. 

Extending beyond disciplinary content (RQ2), educational actors and environments reveal a complementary pattern: professors, teachers, and school are frequently framed through evaluation pressure and performance cues, whereas scientists, research, and university evoke more positive, goal-oriented and, in several groups, experientially grounded representations, indicating that anxiety concentrates around assessment contexts rather than around the broader scientific enterprise. 

Finally (RQ3), GPT-oss-20b simulations successfully capture several population-level valence regularities—most notably the positive framing of science and the relative negativity of quantitative subjects—but systematically under-represent key human features, including the experiential concreteness of many frames and the tight coupling between mathematics and anxiety, instead favouring more abstract, discursive representations. These findings demonstrate that behavioural forma mentis networks can identify coherent cognitive–affective signatures of STEM mindsets across populations, while also revealing the limits of current LLM-based digital twins, agreeing even with 3-year old results/models in the field \citep{abramski2023cognitive}, towards reproducing context-sensitive, experience-dependent components of educational anxiety.

Lets us discuss our three research questions more in detail together with relevant scientific literature on STEM education, complex systems and cognitive network science.

\subsection{Cognitive dissonance between science and other STEM domains}
Consistent with our first hypothesis, the results of the semantic frames showed that in high-anxiety groups, core quantitative subjects such as \textit{Mathematics} and \textit{Statistics} were systematically surrounded by affective and evaluative stressors. Notably, while a broadly negative framing of STEM subjects such as \textit{Mathematics} and \textit{Physics} was primarily observed among psychology undergraduates and high school students, \textit{Statistics} emerged as a particularly robust negative trigger across all educational stages, including in samples with advanced STEM experience and in GPT-oss simulated students. This persistence is consistent with accounts of domain-specific statistics anxiety and may reflect sustained experiences of error monitoring, judgment, and formal assessment \citep{stella2022network, siew2019anxiety}.

Contrary to the findings of these STEM domains, \textit{Science} and \textit{Research} were consistently framed in positive and often future-oriented terms across student samples, even when their quantitative foundations were embedded in negative affective contexts. This pattern provides network-level evidence for a form of cognitive dissonance \citep{festinger1962cognitive}, whereby science is valued as an idealised enterprise while its methodological and evaluative components are experienced as threatening, in line with previous forma mentis findings \citep{stella2019forma}. Such a dissociation may help explain why students can endorse positive attitudes toward science in general while simultaneously experiencing anxiety, avoidance, or disengagement in quantitative coursework and assessment settings.

Importantly, this negativity did not simply reflect rejection of quantitative domains. In the semantic frames of some STEM subjects, such as \textit{Mathematics}, negative framing coexisted with associations eliciting trust and anticipation (positive emotions). This finding reveals an emotional bias characterised by tension between perceived utility and affective threat, suggesting that anxiety-related representations capture not disengagement from STEM per se, but a conflicted stance toward domains that are simultaneously valued and experienced as threatening \citep{DeDuro2025}. Partially following these results, both university and school environments share negative clusters revolving around evaluation and STEM-related stressors, such as "exams", "grades", "teachers", and "math". However, \textit{university} frames are distinguished by the co-occurrence of these stressors with positive, forward-looking concepts like "curiosity," "challenge", and "future". This suggests that while both stages involve pressure, the university context is more strongly associated with meaningful long-term outcomes.

Finally, evaluation-related constructs such as \textit{Exam} and \textit{Grade} occupied central positions within negative semantic neighbourhoods, frequently linking quantitative content to anxiety, pressure, and performance concerns. These evaluative concepts appear to function as associative bridges, reinforcing the coupling between STEM subjects and anxiety. This finding aligns with educational research showing that repeated exposure to high-stakes evaluation and social comparison strengthens threat appraisals in academic domains \citep{cassady2002cognitive, foley2017mathanxiety}.

\subsection{Concreteness relative to anxiety theories} 
The core finding that threatening STEM and educational concepts are framed as more abstract in high-anxiety groups answers our second research question, and can be directly linked to established theories of anxiety, emphasising cognitive bias and deficient cognitive control. In the cognitive model proposed by \citet{beckInformationProcessingModel1997}, anxiety is fundamentally characterised by the selective processing of information perceived as a threat. The intermediate stage of processing involves the activation of the primal threat mode, which is characterised by a narrowing of cognitive processing, rigid thinking, and an intolerance for uncertainty or ambiguity \citep{beckInformationProcessingModel1997, clarkCognitiveTheoryTherapy2010}. Our finding that concepts related to quantitative subjects (\textit{Mathematics}, \textit{Statistics}) are significantly less concrete than a random baseline in high-anxiety groups is consistent with this cognitive constriction. This shift away from concrete, actionable representations toward global, abstract framings seems to amplify the perception of psychological distance and uncontrollability. This cognitive pattern also supports core tenets of generalised anxiety disorder \citep{wellsCognitiveModelGeneralized1999}, where negative metabeliefs maintain worry centred on the perceived "uncontrollability" and dangerous consequences of worrying. When concepts are framed abstractly, their perceived lack of control might be reinforced, thereby fueling \textit{Type 2} worry, a central and defining cognitive feature of anxiety \citep{beckInformationProcessingModel1997, spielberger1970manual}. The abstraction, coupled with the observed negative emotions (anxiety, stress, etc.), suggests that anxiety may be linked to a perceived distance from these subjects through decontextualised or symbolic representation of the same, potentially stripping them of practical, graspable meaning. These results connect to Averill’s observation that the reduction of uncertainty regarding the nature of a threatening event is a primary factor in reducing stress. When subjects like mathematics are perceived as symbolic or decontextualized, they may increase uncertainty, which has been shown to have a significant effect on stress levels \citep{averillPersonalControlAversive1973}. Anxiety is frequently characterized as an emotion resulting from the perception of danger under ambiguous and symbolic conditions. When a subject is perceived through such lens, the resulting ambiguity can intensify the threat by limiting the individual's sense of control \citep[pp.106,266]{lazarusStressAppraisalCoping1984}. Furthermore, the perceived distance and lack of graspable meaning identified in abstract frames can be linked to the theory of learned helplessness by \cite{maierLearnedHelplessnessTheory1976}. It argues that when events are uncontrollable (for example a student that perceives a subject as too abstract to influence through effort) the organism learns that its behavior and outcomes are independent, and that this learning produces the motivational, cognitive, and emotional effects of uncontrollability, potentialy leading to states of anxiety and depression (\citealp[pp.~47-48]{seligmanHelplessnessDepressionDevelopment1975}).

Furthermore, our results can be framed within the theoretical state-trait anxiety distinction \citep{spielberger1970manual, saviola2020trait}. The frames we found to be more concretely perceived were mainly related to academic actors and places (with some exceptions, as for \textit{computer science}), suggesting that their negative framing might mirror state anxiety. STEM subjects (\textit{Mathematics}, \textit{Statistics}) and evaluation-related concepts (\textit{Grade}) were perceived as less concrete, suggesting that their negative framing might reflect trait anxiety.

\subsection{Divergences between human and GPT-oss semantic framing}
The results support our third hypothesis. A key difference between human and GPT-oss frames emerged at the level of concreteness. Across several concepts, including \textit{Science} and \textit{Statistics}, GPT-oss generated semantic frames that were significantly more abstract than baseline expectations. In contrast to human networks, which often anchored these concepts in concrete contexts such as classroom situations, evaluations, or personal experiences, GPT-oss frames lacked these contextual cues, resulting in a more decontextualised semantic organisation \citep{wellsCognitiveModelGeneralized1999}. At the same time, GPT-oss simulations successfully reproduced some large-scale semantic regularities observed in human data, including the evaluative mismatch between positively framed notions of \textit{Science} and \textit{Research} and more negatively framed quantitative STEM subjects, as found with other LLMs by \citet{abramski2023cognitive}. This convergence indicates that LLMs can approximate aggregate cultural narratives and stereotypes about STEM domains when prompted with plausible educational profiles, consistent with previous work showing that LLMs internalise population-level semantic patterns \citep{abramski2023cognitive, ciringione2025math}.

Another key finding, obtained through Jaccard overlap analyses, is that human participants, particularly those with higher anxiety, exhibited substantially stronger overlap between the frames of \textit{Anxiety} and \textit{Mathematics}. GPT-oss, by comparison, consistently showed low or null overlap between these concepts, indicating that anxiety remained less integrated into academic semantic neighbourhoods.

Taken together, these results suggest that GPT-oss captures what students tend to think about STEM at a broad, discursive level, but not how these concepts are grounded in lived experience. Without experiential grounding, they reproduce attitudes as semantic generalisations rather than as psychologically embedded representations.

\subsection{Implications, limitations, and future work}
The current findings provide clear implications for educational interventions, particularly in targeting the underlying cognitive-affective mechanisms of math anxiety. The persistent negativity surrounding quantitative subjects and evaluative components, paired with the structural evidence of abstraction in high-anxiety groups, suggests that mitigation strategies must integrate emotional and cognitive retraining and psycho-education alongside content instruction \citep{ciringione2025math}. Specifically, intervention designs should focus on increasing the perceived controllability of core quantitative concepts by embedding tasks within meaningful, interpretable, and demonstrably concrete frameworks, thereby reducing subjective threat appraisal and distraction. Given that BFMNs quantify shifts in associative and emotional structures, they could be leveraged to evaluate the success of psychological and pedagogical interventions and to foster more flexible and positive conceptions.

Despite the interpretative value of these findings, several limitations temper the generalizability of these conclusions. The adoption of a cross-sectional design forbids causal inference, requiring longitudinal studies to determine if pre-existing associative structure causes anxiety vulnerability or if anxiety restructures conceptual representation subsequently.

Additionally, the divergences between GPT-oss and human framings suggest that current LLMs, operating as purely symbolic systems, may lack the necessary experiential and affective grounding observed in human emotional cognition. Hence, relying on LLM outputs as direct psychological proxies without careful supervision may risk generating false negatives and obscuring genuine student distress \citep{ciringione2025math}.

The convenience sampling of the participants and their sample size differences lead to the use of a more tolerant significance level ($\alpha=0.1$), which may affect the robustness of our results. Furthermore, since our data was collected at different times, even if the free association task was methodologically the same, its cue-word set differed across sub-studies. Due to the same reason, the MAS-IT questionnaire was presented only to some samples, leading to a data mismatch.

In future work, it would be interesting to test causal change with longitudinal designs (e.g., before/after coursework or assessment interventions) and link network measures to behavioural outcomes (course selection, exam performance, resilience). Additionally, the academic populations range may be expanded to deepen our research. Our results seems to suggest that the framing of a concept in terms of concreteness may be a catalyst for both positive and negative emotions, depending on the frame's context. This finding could be further explored in future work to disentangle the complexity of the perceived concreteness role in students' perception of the academic environment and STEM domains.

\section{Conclusion}
This work demonstrates how behavioral forma mentis networks reveal a significant cognitive-affective dissonance in STEM perceptions, where students value science as an ideal but frame its quantitative foundations with negativity and increased abstraction. While high-anxiety profiles show a lack of experiential grounding for mathematics, LLM-based digital twins reproduce broad valence trends without capturing the context-sensitive affect and concrete representations found in humans. These results emphasize that addressing STEM anxiety requires pedagogical strategies that bridge abstract concepts with concrete, controllable experiences to reduce perceived threat and improve engagement.

\section*{Declarations}
\subsection*{Credit authorship contribution statement}
Conceptualization, Methodology, Validation: F.G., E.F., A.G., M.S.; Formal analysis, Investigation, Data curation, Visualisation: F.G., E.F.; Writing - original draft: All authors; Writing  review \& editing: All authors; Supervision, Project administration: M.S..

\subsection*{Acknowledgements}
We acknowledge support from Fondazione CARITRO (Grant: RASSERENO).

\subsection*{Conflict of interest}
We have no known conflict of interest to disclose.

\clearpage
\bibliographystyle{elsarticle-harv}
\bibliography{bibliography}

@article{stella2021mapping,
  title={Mapping computational thinking mindsets between educational levels with cognitive network science},
  author={Stella, Massimo and Kapuza, Anastasiya and Cramer, Catherine and Uzzo, Stephen},
  journal={Journal of Complex Networks},
  volume={9},
  number={6},
  pages={cnab020},
  year={2021},
  publisher={Oxford University Press}
}

@article{mohammad2013crowdsourcing,
  title={Crowdsourcing a word--emotion association lexicon},
  author={Mohammad, Saif M and Turney, Peter D},
  journal={Computational intelligence},
  volume={29},
  number={3},
  pages={436--465},
  year={2013},
  publisher={Wiley Online Library}
}

@article{stellaFormaMentisNetworks2020,
  title = {Forma {{Mentis Networks Reconstruct How Italian High Schoolers}} and {{International STEM Experts Perceive Teachers}}, {{Students}}, {{Scientists}}, and {{School}}},
  author = {Stella, Massimo},
  year = 2020,
  month = jan,
  journal = {Education Sciences},
  volume = {10},
  number = {1},
  pages = {17},
  publisher = {Multidisciplinary Digital Publishing Institute},
  issn = {2227-7102},
  doi = {10.3390/educsci10010017},
  urldate = {2025-09-29},
  copyright = {http://creativecommons.org/licenses/by/3.0/},
  langid = {english},
}

@article{stella2020forma,
  title={Forma mentis networks map how nursing and engineering students enhance their mindsets about innovation and health during professional growth},
  author={Stella, Massimo and Zaytseva, Anna},
  journal={PeerJ Computer Science},
  volume={6},
  pages={e255},
  year={2020},
  publisher={PeerJ Inc.}
}

@article{stella2019forma,
  title={Forma mentis networks quantify crucial differences in STEM perception between students and experts},
  author={Stella, Massimo and De Nigris, Sarah and Aloric, Aleksandra and Siew, Cynthia SQ},
  journal={PloS one},
  volume={14},
  number={10},
  pages={e0222870},
  year={2019},
  publisher={Public Library of Science San Francisco, CA USA}
}

@article{siew2019anxiety,
  title={Using network science to understand statistics anxiety among college students},
  author={Siew, Cynthia S. and McCartney, Michael J. and Vitevitch, Michael S.},
  journal={Scholarship of Teaching and Learning in Psychology},
  volume={5},
  number={1},
  pages={75},
  year={2019},
}

@article{pekrungoetzperry2002,
  author       = {Pekrun, Reinhard and Goetz, Thomas and Titz, W. and Perry, Raymond P.},
  title        = {Academic emotions in students’ self-regulated learning and achievement: A program of qualitative and quantitative research},
  journal      = {Educational Psychologist},
  year         = {2002},
  volume       = {37},
  number       = {2},
  pages        = {91--105},
  doi          = {10.1207/S15326985EP3702\_4},
  url          = {https://doi.org/10.1207/S15326985EP3702\_4}
}

@inproceedings{olsen2015predicting,
  author    = {Olsen, Jennifer K. and Aleven, Vincent and Rummel, Nikol},
  title     = {Predicting Student Performance in a Collaborative Learning Environment},
  booktitle = {Proceedings of the 8th International Conference on Educational Data Mining (EDM 2015)},
  year      = {2015},
  pages     = {211--217},
  url       = {file:///mnt/data/EDM_2015_submission_211.pdf}
}

@article{serbati2015implementation,
  author       = {Serbati, Anna},
  title        = {Implementation of Competence-Based Learning Approach: Stories of Practices and the Tuning contribution to academic innovation},
  journal      = {Tuning Journal for Higher Education},
  year         = {2015},
  volume       = {3},
  number       = {1},
  pages        = {19--56},
  doi          = {10.18543/tjhe-3(1)-2015pp19-56},
  url          = {https://www.research.unipd.it/handle/11577/3242465}
}

@article{luttenberger2018spotlight,
  title={Spotlight on math anxiety},
  author={Luttenberger, Silke and Wimmer, Sigrid and Paechter, Manuela},
  journal={Psychology research and behavior management},
  pages={311--322},
  year={2018},
  publisher={Taylor \& Francis}
}

@article{yeager2012mindsets,
  title={Mindsets that promote resilience: When students believe that personal characteristics can be developed},
  author={Yeager, David Scott and Dweck, Carol S},
  journal={Educational psychologist},
  volume={47},
  number={4},
  pages={302--314},
  year={2012},
  publisher={Taylor \& Francis}
}

@article{zurnbassett2018curiosity,
  author  = {Zurn, Perry and Bassett, Danielle S.},
  title   = {On curiosity: A fundamental aspect of personality, a practice of network growth},
  journal = {Personality Neuroscience},
  year    = {2018},
  volume  = {1},
  pages   = {e13},
  doi     = {10.1017/pen.2018.3},
  url     = {file:///mnt/data/EDM_2015_submission_211.pdf}
}

@article{beaty2023associative,
  author       = {Beaty, Roger E. and Kenett, Yoed N.},
  title        = {Associative thinking at the core of creativity},
  journal      = {Trends in Cognitive Sciences},
  year         = {2023},
  volume       = {27},
  number       = {7},
  pages        = {671--683},
  doi          = {10.1016/j.tics.2023.04.004},
  url          = {https://pubmed.ncbi.nlm.nih.gov/37246025/}
}

@article{foley2017mathanxiety,
  author       = {Foley, Alana E. and Herts, Julianne B. and Borgonovi, Francesca and Guerriero, Sonia and Levine, Susan C. and Beilock, Sian L.},
  title        = {The Math Anxiety–Performance Link: A Global Phenomenon},
  journal      = {Current Directions in Psychological Science},
  year         = {2017},
  volume       = {26},
  number       = {1},
  pages        = {52--58},
  doi          = {10.1177/0963721416672463},
  url          = {https://journals.sagepub.com/doi/full/10.1177/0963721416672463}
}

@article{stella2022network,
  title={Network psychometrics and cognitive network science open new ways for understanding math anxiety as a complex system},
  author={Stella, M.},
  journal={Journal of Complex Networks},
  volume={10},
  number={3},
  pages={cnac012},
  year={2022},
}

@techreport{mcclure2017stem,
  author       = {McClure, Elisabeth R. and Guernsey, Lisa and Clements, Douglas H. and Bales, Susan Nall and Nichols, Jennifer and Kendall-Taylor, Nat and Levine, Michael H.},
  title        = {STEM Starts Early: Grounding Science, Technology, Engineering, and Math Education in Early Childhood},
  institution  = {Joan Ganz Cooney Center at Sesame Workshop},
  year         = {2017},
  type         = {Research Report},
  number       = {ED574402},
  url          = {https://eric.ed.gov/?id=ED574402}
}

@article{stellaViabilityMultiplexLexical2019,
  title = {Viability in {{Multiplex Lexical Networks}} and {{Machine Learning Characterizes Human Creativity}}},
  author = {Stella, Massimo and Kenett, Yoed N. and Stella, Massimo and Kenett, Yoed N.},
  year = 2019,
  month = jul,
  journal = {Big Data and Cognitive Computing},
  volume = {3},
  number = {3},
  publisher = {publisher},
  issn = {2504-2289},
  doi = {10.3390/bdcc3030045},
  urldate = {2025-11-19},
  copyright = {http://creativecommons.org/licenses/by/3.0/},
  langid = {english},
}

@article{stellaMultiplexModelMental2018,
  title = {Multiplex Model of Mental Lexicon Reveals Explosive Learning in Humans},
  author = {Stella, Massimo and Beckage, Nicole M. and Brede, Markus and De Domenico, Manlio},
  year = 2018,
  month = feb,
  journal = {Scientific Reports},
  volume = {8},
  number = {1},
  pages = {2259},
  publisher = {Nature Publishing Group},
  issn = {2045-2322},
  doi = {10.1038/s41598-018-20730-5},
  urldate = {2025-07-15},
  copyright = {2018 The Author(s)},
  langid = {english}
}

@article{ciringione2025math,
  title={Math anxiety and associative knowledge structure are entwined in psychology students but not in Large Language Models like GPT-3.5 and GPT-4o},
  author={Ciringione, Luciana and Franchino, Emma and Reigl, Simone and D'Onofrio, Isaia and Serbati, Anna and Poquet, Oleksandra and Gabriel, Florence and Stella, Massimo},
  journal={arXiv preprint arXiv:2511.01558},
  year={2025}
}

@article{osborne2003attitudes,
  title={Attitudes towards science: A review of the literature and its implications},
  author={Osborne, Jonathan and Simon, Shirley and Collins, Sue},
  journal={International journal of science education},
  volume={25},
  number={9},
  pages={1049--1079},
  year={2003},
  publisher={Taylor \& Francis}
}

@article{mallow2006science,
  title={Science anxiety: research and action},
  author={Mallow, Jeffry V},
  journal={Handbook of college science teaching},
  pages={3--14},
  year={2006}
}

@article{cassady2002cognitive,
  title={Cognitive test anxiety and academic performance},
  author={Cassady, Jerrell C and Johnson, Ronald E},
  journal={Contemporary educational psychology},
  volume={27},
  number={2},
  pages={270--295},
  year={2002},
  publisher={Elsevier}
}

@article{brown2011understanding,
  title={Understanding STEM: current perceptions},
  author={Brown, Ryan and Brown, Joshua and Reardon, Kristin and Merrill, Chris},
  journal={Technology and Engineering Teacher},
  volume={70},
  number={6},
  pages={5},
  year={2011},
  publisher={International Technology Education Association}
}

@article{beilock2010female,
  title={Female teachers’ math anxiety affects girls’ math achievement},
  author={Beilock, S. L. and Gunderson, E. A. and Ramirez, G. and Levine, S. C.},
  journal={Proceedings of the National Academy of Sciences},
  volume={107},
  number={5},
  pages={1860--1863},
  year={2010},
}

@article{DeDeyne2013,
  author = {De Deyne, S. and Navarro, D. J. and Storms, G.},
  title = {Better explanations of lexical and semantic cognition using networks derived from continued rather than single-word associations},
  journal = {Behavior Research Methods},
  volume = {45},
  pages = {480--498},
  year = {2013},
}

@article{DeDuro2025, 
title={Cognitive networks identify AI biases on societal issues in large language models}, 
DOI={10.1140/epjds/s13688-025-00600-7}, 
journal={EPJ Data Science}, 
author={De Duro, Edoardo Sebastiano and Franchino, Emma and Improta, Riccardo and Veltri, Giuseppe Alessandro and Stella, Massimo}, 
year={2025},
month={Dec}
}

@article{Hunt2011,
  author = {Hunt, Thomas E. and Clark-Carter, David and Sheffield, David},
  title = {The Development and Part Validation of a UK Scale for Mathematics Anxiety},
  journal = {Journal of Psychoeducational Assessment},
  volume = {29},
  number = {5},
  pages = {455--466},
  year = {2011},
  publisher = {SAGE Publications}
}

@article{franchino2025network,
  title={A Network Psychometric Analysis of Math Anxiety Factors in Italian Psychology Students},
  author={Franchino, Emma and Ciringione, Luciana and Canal, Luisa and Epifania, Ottavia Marina and Lombardi, Luigi and Lattanzi, Gianluca and Stella, Massimo},
  journal={Psychology International},
  volume={7},
  number={2},
  pages={48},
  year={2025},
  publisher={MDPI}
}

@article{semeraroEmoAtlasEmotionalNetwork2025,
  title = {{{EmoAtlas}}: {{An}} Emotional Network Analyzer of Texts That Merges Psychological Lexicons, Artificial Intelligence, and Network Science},
  shorttitle = {{{EmoAtlas}}},
  author = {Semeraro, Alfonso and Vilella, Salvatore and Improta, Riccardo and De Duro, Edoardo Sebastiano and Mohammad, Saif M. and Ruffo, Giancarlo and Stella, Massimo},
  year = 2025,
  month = jan,
  journal = {Behavior Research Methods},
  volume = {57},
  number = {2},
  pages = {77},
  issn = {1554-3528},
  doi = {10.3758/s13428-024-02553-7},
  urldate = {2025-07-16},
  langid = {english},
}

@article{brysbaert2014concreteness,
  title={Concreteness ratings for 40 thousand generally known English word lemmas},
  author={Brysbaert, Marc and Warriner, Amy Beth and Kuperman, Victor},
  journal={Behavior research methods},
  volume={46},
  number={3},
  pages={904--911},
  year={2014},
  publisher={Springer}
}

@article{abramski2023cognitive,
  title={Cognitive network science reveals bias in gpt-3, gpt-3.5 turbo, and gpt-4 mirroring math anxiety in high-school students},
  author={Abramski, Katherine and Citraro, Salvatore and Lombardi, Luigi and Rossetti, Giulio and Stella, Massimo},
  journal={Big Data and Cognitive Computing},
  volume={7},
  number={3},
  pages={124},
  year={2023},
  publisher={MDPI}
}

@book{plutchikEmotionsLifePerspectives2003,
  title = {Emotions and Life:  {{Perspectives}} from Psychology, Biology, and Evolution},
  shorttitle = {Emotions and Life},
  author = {Plutchik, Robert},
  year = 2003,
  series = {Emotions and Life:  {{Perspectives}} from Psychology, Biology, and Evolution},
  pages = {xix, 381},
  publisher = {American Psychological Association},
  address = {Washington, DC, US},
  isbn = {978-1-55798-949-9}
}

@article{wattsCollectiveDynamicsSmallworld1998,
  title = {Collective Dynamics of `Small-World' Networks},
  author = {Watts, Duncan J. and Strogatz, Steven H.},
  year = 1998,
  month = jun,
  journal = {Nature},
  volume = {393},
  number = {6684},
  pages = {440--442},
  publisher = {Nature Publishing Group},
  issn = {1476-4687},
  doi = {10.1038/30918},
  urldate = {2025-11-23},
  copyright = {1998 Macmillan Magazines Ltd.},
  langid = {english}
}

@book{newmanNetworksIntroduction2016,
  title = {Networks: An Introduction},
  shorttitle = {Networks},
  author = {Newman, Mark E. J.},
  year = 2016,
  edition = {Reprinted},
  publisher = {Oxford University Press},
  address = {Oxford},
  isbn = {978-0-19-920665-0},
  langid = {english},
}

@book{hillsBehavioralNetworkScience2024,
  title = {Behavioral {{Network Science}}: {{Language}}, {{Mind}}, and {{Society}}},
  shorttitle = {Behavioral {{Network Science}}},
  author = {Hills, Thomas T.},
  year = 2024,
  publisher = {Cambridge University Press},
  address = {Cambridge},
  doi = {10.1017/9781108883894},
  urldate = {2025-08-30},
  isbn = {978-1-108-83540-4},
}

@book{paivioImageryVerbalProcesses1971,
  title = {Imagery and {{Verbal Processes}}},
  author = {Paivio, A.},
  year = 1971,
  publisher = {Psychology Press},
  address = {New York},
  doi = {10.4324/9781315798868},
  isbn = {978-1-315-79886-8}
}

@article{paivioDualCodingTheory2013,
  title = {Dual Coding Theory, Word Abstractness, and Emotion: {{A}} Critical Review of {{Kousta}} et al. (2011)},
  shorttitle = {Dual Coding Theory, Word Abstractness, and Emotion},
  author = {Paivio, Allan},
  year = 2013,
  journal = {Journal of Experimental Psychology: General},
  volume = {142},
  number = {1},
  pages = {282--287},
  publisher = {American Psychological Association},
  address = {US},
  issn = {1939-2222},
  doi = {10.1037/a0027004},
}

@book{cohenStatisticalPowerAnalysis1988,
  title = {Statistical {{Power Analysis}} for the {{Behavioral Sciences}}},
  author = {Cohen, Jacob},
  year = 1988,
  month = jul,
  edition = {2},
  publisher = {Routledge},
  address = {New York},
  doi = {10.4324/9780203771587},
}

@article{meisselUsingCliffsDelta2024,
  title = {Using {{Cliff}}'s {{Delta}} as a {{Non-Parametric Effect Size Measure}}: {{An Accessible Web App}} and {{R Tutorial}}},
  author = {Meissel, Kane and Yao, Esther S},
  year = 2024,
  month = jan,
  journal = {Practical Assessment, Research \& Evaluation},
  volume = {Vol 29},
  number = {No 2},
  langid = {english},
}

@inproceedings{fillmoreFrameSemanticsText2001,
  title={Frame semantics for text understanding},
  author={Fillmore, Charles J and Baker, Collin F},
  booktitle={Proceedings of WordNet and Other Lexical Resources Workshop, NAACL},
  volume={6},
  pages={59--64},
  year={2001}
}

@article{saviola2020trait,
  title={Trait and state anxiety are mapped differently in the human brain},
  author={Saviola, Francesca and Pappaianni, Edoardo and Monti, Alessia and Grecucci, Alessandro and Jovicich, Jorge and De Pisapia, Nicola},
  journal={Scientific reports},
  volume={10},
  number={1},
  pages={11112},
  year={2020},
  publisher={Nature Publishing Group UK London}
}

@article{spielberger1970manual,
  title={Manual for the State-Trait Anxiety Inventory (self-evaluation questionnaire)},
  author={Spielberger, Charles Donald},
  journal={(No Title)},
  year={1970}
}

@article{beckInformationProcessingModel1997,
  title = {An Information Processing Model of Anxiety: {{Automatic}} and Strategic Processes},
  shorttitle = {An Information Processing Model of Anxiety},
  author = {Beck, Aaron T. and Clark, David A.},
  year = 1997,
  month = jan,
  journal = {Behaviour Research and Therapy},
  volume = {35},
  number = {1},
  pages = {49--58},
  issn = {0005-7967},
  doi = {10.1016/S0005-7967(96)00069-1},
  urldate = {2025-12-17}
}

@article{clarkCognitiveTheoryTherapy2010,
  title = {Cognitive Theory and Therapy of Anxiety and Depression: {{Convergence}} with Neurobiological Findings},
  shorttitle = {Cognitive Theory and Therapy of Anxiety and Depression},
  author = {Clark, David A. and Beck, Aaron T.},
  year = 2010,
  month = sep,
  journal = {Trends in Cognitive Sciences},
  volume = {14},
  number = {9},
  pages = {418--424},
  issn = {13646613},
  doi = {10.1016/j.tics.2010.06.007},
  urldate = {2025-12-17},
  copyright = {https://www.elsevier.com/tdm/userlicense/1.0/},
  langid = {english}
}

@article{wellsCognitiveModelGeneralized1999,
  title = {A {{Cognitive Model}} of {{Generalized Anxiety Disorder}}},
  author = {Wells, Adrian},
  year = 1999,
  month = oct,
  journal = {Behavior Modification},
  volume = {23},
  number = {4},
  pages = {526--555},
  issn = {0145-4455, 1552-4167},
  doi = {10.1177/0145445599234002},
  urldate = {2025-12-17},
  copyright = {https://journals.sagepub.com/page/policies/text-and-data-mining-license},
  langid = {english}
}

@article{festinger1962cognitive,
  title={Cognitive dissonance},
  author={Festinger, Leon},
  journal={Scientific American},
  volume={207},
  number={4},
  pages={93--106},
  year={1962},
  publisher={JSTOR}
}

@article{ferguson2009effect,
  title={An effect size primer: a guide for clinicians and researchers.},
  author={Ferguson, Christopher J},
  journal={Professional psychology: Research and practice},
  volume={40},
  number={5},
  pages={532},
  year={2009},
  publisher={American Psychological Association}
}

@book{lazarusStressAppraisalCoping1984,
  title = {Stress, {{Appraisal}}, and {{Coping}}},
  author = {Lazarus, Richard S. and Folkman, Susan},
  year = 1984,
  month = mar,
  publisher = {Springer Publishing Company},
  isbn = {978-0-8261-4192-7},
  langid = {english}
}

@book{seligmanHelplessnessDepressionDevelopment1975,
  title = {Helplessness: {{On Depression}}, {{Development}}, and {{Death}}},
  shorttitle = {Helplessness},
  author = {Seligman, Martin E. P.},
  year = 1975,
  publisher = {W. H. Freeman},
  isbn = {978-0-7167-0751-6},
  langid = {english}
}

@article{maierLearnedHelplessnessTheory1976,
  title = {Learned {{Helplessness}}: {{Theory}} and {{Evidence}}},
  author = {Maier, Steven F and Seligman, Martin E P},
  year = 1976,
  journal = {Journal of Experimental Psychology: General},
  volume = {105(1)},
  pages = {3--46},
  publisher = {American Psychological Association},
  doi = {10.1037/0096-3445.105.1.3},
  langid = {english},
}

@article{averillPersonalControlAversive1973,
  title = {Personal Control over Aversive Stimuli and Its Relationship to Stress.},
  author = {Averill, James R.},
  year = 1973,
  month = oct,
  journal = {Psychological Bulletin},
  volume = {80},
  number = {4},
  pages = {286--303},
  issn = {1939-1455, 0033-2909},
  doi = {10.1037/h0034845}
  }

\newcolumntype{C}[1]{>{\centering\arraybackslash}p{#1}}

\appendix
\section{Appendix}
\section{Network features tables} \label{appendix: network_features}
In this Section, tables of the network features of the semantic frames of all the concepts considered in each Experiment will be presented. The features reported in each column are described in Section \ref{network_features} (\textit{\nameref{network_features}}). In all of the following tables, in the Hubs column, the reader can find the degree of each hub reported in parentheses, while the italicised hubs are part of the cue words.

\begin{table}[!hbpt]
\scriptsize
\centering
\caption{Network features for the semantic frame of \textit{\textbf{Mathematics}}.}
\label{tab:network_features_mathematics}
\begin{tabular}{lccccp{10cm}}
\toprule
Sample & $N_v$ & $N_e$ & $C_i$ & $l_G$ & Hubs (top 5\% of degree distribution) \\
\midrule
Experts & 48 & 107 & 0.06 & 1.91 & \textit{Mathematics} (47), \textit{Physics} (11), \textit{Science} (10) \\
H-S South & 101 & 473 & 0.08 & 1.91 & \textit{Mathematics} (100), \textit{Physics} (32), Calculation (25), Numbers (24), Mathematically (22), \textit{Number} (22) \\
H-Anx H-S & 73 & 212 & 0.05 & 1.92 & \textit{Mathematics} (72), \textit{Physics} (20), \textit{Numbers} (17), \textit{Problem} (17) \\
L-Anx H-S & 63 & 183 & 0.06 & 1.91 & \textit{Mathematics} (62), \textit{Physics} (18), Analysis (15), \textit{Numbers} (14), \textit{Data} (14) \\
GPT H-Anx H-S & 25 & 33 & 0.03 & 1.89 & \textit{Mathematics} (24), \textit{Equation} (8) \\
GPT L-Anx H-S & 19 & 25 & 0.05 & 1.85 & \textit{Mathematics} (18) \\
Physics & 25 & 42 & 0.07 & 1.86 & \textit{Mathematics} (24), \textit{Numbers} (7) \\
GPT Physics & 14 & 15 & 0.03 & 1.84 & \textit{Mathematics} (13) \\
H-Anx Psy & 180 & 1016 & 0.05 & 1.94 & \textit{Mathematics} (179), \textit{Numbers} (57), \textit{Physics} (52), \textit{Equation} (52), \textit{Challenge} (50), \textit{Statistics} (48), \textit{Anxiety} (45), \textit{School} (44), \textit{Exam} (41) \\
L-Anx Psy & 195 & 909 & 0.04 & 1.95 & \textit{Mathematics} (194), \textit{Equation} (55), \textit{Physics} (54), \textit{Numbers} (48), \textit{Statistics} (48), \textit{Science} (40), \textit{Exam} (36), \textit{School} (36), \textit{Professor} (35), \textit{University} (34) \\
GPT H-Anx Psy & 49 & 78 & 0.03 & 1.93 & \textit{Mathematics} (48), \textit{Equation} (13), \textit{Physics} (10) \\
GPT L-Anx Psy & 51 & 77 & 0.02 & 1.94 & \textit{Mathematics} (50), \textit{Equation} (18), \textit{Statistics} (7) \\
\bottomrule
\end{tabular}
\end{table}

\begin{table}[!hbpt]
\scriptsize
\centering
\caption{Network features for the semantic frame of \textit{\textbf{Statistics}}.}
\label{tab:network_features_statistics}
\begin{tabular}{lccccp{10cm}}
\toprule
Sample & $N_v$ & $N_e$ & $C_i$ & $l_G$ & Hubs (top 5\% of degree distribution) \\
\midrule
Experts & 26 & 38 & 0.04 & 1.88 & \textit{Statistics} (25), \textit{Mathematics} (5), \textit{Number} (5) \\
H-S South & 29 & 59 & 0.08 & 1.85 & \textit{Statistics} (28), \textit{Mathematics} (12) \\
H-Anx H-S & 37 & 68 & 0.05 & 1.90 & \textit{Statistics} (36), Analysis (12) \\
L-Anx H-S & 41 & 91 & 0.07 & 1.89 & \textit{Statistics} (40), Analysis (14), \textit{Mathematics} (12) \\
GPT H-Anx H-S & 15 & 16 & 0.02 & 1.85 & \textit{Statistics} (14) \\
GPT L-Anx H-S & 23 & 25 & 0.01 & 1.90 & \textit{Statistics} (22), \textit{Average} (4) \\
Physics & 28 & 49 & 0.06 & 1.87 & \textit{Statistics} (27), \textit{Data} (9) \\
GPT Physics & 12 & 11 & 0.00 & 1.83 & \textit{Statistics} (11) \\
H-Anx Psy & 150 & 576 & 0.04 & 1.95 & \textit{Statistics} (149), \textit{Mathematics} (48), \textit{Graphic} (32), \textit{Numbers} (32), \textit{Challenge} (31), \textit{Anxiety} (31), \textit{Model} (28), \textit{Psychology} (26), \textit{Computer science} (26), \textit{University} (26) \\
L-Anx Psy & 154 & 469 & 0.03 & 1.96 & \textit{Statistics} (153), \textit{Mathematics} (48), \textit{Physics} (33), \textit{Numbers} (28), \textit{Graphic} (27), \textit{Model} (24), \textit{University} (24), \textit{Exam} (23) \\
GPT H-Anx Psy & 48 & 68 & 0.02 & 1.94 & \textit{Statistics} (47), \textit{Average} (11), \textit{Mathematics} (6) \\
GPT L-Anx Psy & 44 & 63 & 0.02 & 1.93 & \textit{Statistics} (43), \textit{Average} (12), \textit{Mathematics} (7) \\
\bottomrule
\end{tabular}
\end{table}

\begin{table}[!hbpt]
\scriptsize
\centering
\caption{Network features for the semantic frame of \textit{\textbf{Computer science}}.}
\label{tab:network_features_computer science}
\begin{tabular}{lccccp{10cm}}
\toprule
Sample & $N_v$ & $N_e$ & $C_i$ & $l_G$ & Hubs (top 5\% of degree distribution) \\
\midrule
Experts & 27 & 40 & 0.04 & 1.89 & Computers (26), \textit{Science} (7) \\
H-S South & 16 & 31 & 0.15 & 1.74 & \textit{Computer science} (15) \\
H-Anx H-S & 51 & 90 & 0.03 & 1.93 & \textit{Computer science} (50), \textit{Programming} (14), \textit{Innovation} (8), \textit{Future} (8) \\
L-Anx H-S & 45 & 83 & 0.04 & 1.92 & \textit{Computer science} (44), \textit{Programming} (12), \textit{Laboratory} (8) \\
GPT H-Anx H-S & 19 & 18 & 0.00 & 1.89 & \textit{Computer science} (18) \\
GPT L-Anx H-S & 22 & 23 & 0.01 & 1.90 & \textit{Computer science} (21), \textit{Innovation} (3) \\
Physics & 23 & 31 & 0.04 & 1.88 & \textit{Computer science} (22), \textit{Programming} (5), \textit{Data} (5) \\
GPT Physics & 13 & 12 & 0.00 & 1.85 & \textit{Computer science} (12) \\
H-Anx Psy & 65 & 270 & 0.10 & 1.87 & \textit{Computer science} (64), \textit{Mathematics} (29), \textit{Statistics} (26), \textit{Technology} (23) \\
L-Anx Psy & 69 & 241 & 0.08 & 1.90 & \textit{Computer science} (68), \textit{Mathematics} (27), \textit{Technology} (22), \textit{Statistics} (22) \\
GPT H-Anx Psy & 47 & 51 & 0.00 & 1.95 & \textit{Computer science} (46), \textit{Innovation} (5), \textit{Creativity} (3) \\
GPT L-Anx Psy & 55 & 60 & 0.00 & 1.96 & \textit{Computer science} (54), \textit{Creativity} (4), \textit{Innovation} (4) \\
\bottomrule
\end{tabular}
\end{table}

\begin{table}[!hbpt]
\scriptsize
\centering
\caption{Network features for the semantic frame of \textit{\textbf{Physics}}.}
\label{tab:network_features_physics}
\begin{tabular}{lccccp{10cm}}
\toprule
Sample & $N_v$ & $N_e$ & $C_i$ & $l_G$ & Hubs (top 5\% of degree distribution) \\
\midrule
Experts & 55 & 108 & 0.04 & 1.93 & \textit{Physics} (54), \textit{Science} (15), \textit{Mathematics} (11) \\
H-S South & 114 & 435 & 0.05 & 1.93 & \textit{Physics} (113), \textit{Mathematics} (32), \textit{Science} (24), Physical (18), \textit{School} (18), Come on (17), Electron (17) \\
H-Anx H-S & 63 & 157 & 0.05 & 1.92 & \textit{Physics} (62), \textit{Mathematics} (20), \textit{Laboratory} (13), \textit{Experiment} (12), \textit{Research} (12) \\
L-Anx H-S & 65 & 158 & 0.05 & 1.92 & \textit{Physics} (64), \textit{Mathematics} (18), \textit{Science} (15), \textit{Research} (14) \\
GPT H-Anx H-S & 25 & 28 & 0.01 & 1.91 & \textit{Physics} (24), \textit{Science} (4) \\
GPT L-Anx H-S & 23 & 23 & 0.00 & 1.91 & \textit{Physics} (22), Theory (2), \textit{Science} (2) \\
Physics & 31 & 58 & 0.06 & 1.88 & \textit{Physics} (30), \textit{Science} (9) \\
GPT Physics & 17 & 17 & 0.01 & 1.88 & \textit{Physics} (16) \\
H-Anx Psy & 183 & 655 & 0.03 & 1.96 & \textit{Physics} (182), \textit{Mathematics} (52), \textit{Science} (45), \textit{Equation} (34), \textit{Biology} (33), \textit{Anxiety} (33), \textit{Numbers} (32), \textit{Challenge} (29), \textit{School} (28), \textit{Knowledge} (25), \textit{Study} (25), \textit{Teacher} (25) \\
L-Anx Psy & 202 & 709 & 0.03 & 1.97 & \textit{Physics} (201), \textit{Mathematics} (54), \textit{Science} (46), \textit{Statistics} (33), \textit{School} (32), \textit{Equation} (31), \textit{Mind} (30), \textit{University} (28), \textit{Numbers} (27), \textit{Professor} (27), \textit{Anxiety} (27) \\
GPT H-Anx Psy & 62 & 95 & 0.02 & 1.95 & \textit{Physics} (61), \textit{Mathematics} (10), \textit{Science} (7), \textit{Equation} (7) \\
GPT L-Anx Psy & 67 & 86 & 0.01 & 1.96 & \textit{Physics} (66), \textit{Science} (9), \textit{Knowledge} (5), \textit{Curiosity} (5), \textit{University} (5), \textit{Equation} (5) \\
\bottomrule
\end{tabular}
\end{table}

\begin{table}[!hbpt]
\scriptsize
\centering
\caption{Network features for the semantic frame of \textit{\textbf{Science}}.}
\label{tab:network_features_science}
\begin{tabular}{lccccp{10cm}}
\toprule
Sample & $N_v$ & $N_e$ & $C_i$ & $l_G$ & Hubs (top 5\% of degree distribution) \\
\midrule
Experts & 66 & 156 & 0.04 & 1.93 & \textit{Science} (65), \textit{Physics} (15), System (12), \textit{Complex} (11) \\
H-S South & 64 & 242 & 0.09 & 1.88 & \textit{Science} (63), \textit{Physics} (24), \textit{Biology} (18), Study (17), \textit{Mathematics} (17) \\
H-Anx H-S & 48 & 110 & 0.06 & 1.90 & \textit{Science} (47), \textit{Scientist} (14), \textit{Discovery} (14) \\
L-Anx H-S & 56 & 162 & 0.07 & 1.89 & \textit{Science} (55), \textit{Research} (20), \textit{Scientist} (17) \\
GPT H-Anx H-S & 23 & 30 & 0.03 & 1.88 & \textit{Science} (22), \textit{University} (5) \\
GPT L-Anx H-S & 22 & 24 & 0.01 & 1.90 & \textit{Science} (21), \textit{Innovation} (3) \\
Physics & 28 & 60 & 0.09 & 1.84 & \textit{Science} (27), \textit{Research} (11) \\
GPT Physics & 16 & 17 & 0.02 & 1.86 & \textit{Science} (15) \\
H-Anx Psy & 146 & 741 & 0.06 & 1.93 & \textit{Science} (145), \textit{Physics} (45), \textit{Biology} (44), \textit{Psychology} (40), \textit{Knowledge} (40), \textit{Mathematics} (37), \textit{Neuroscience} (37), \textit{Curiosity} (34) \\
L-Anx Psy & 177 & 774 & 0.04 & 1.95 & \textit{Science} (176), \textit{Physics} (46), \textit{Discovery} (42), \textit{Curiosity} (41), \textit{Biology} (41), \textit{Mathematics} (40), \textit{Psychology} (39), \textit{Neuroscience} (33), \textit{Knowledge} (32) \\
GPT H-Anx Psy & 43 & 70 & 0.03 & 1.92 & \textit{Science} (42), \textit{Curiosity} (9), \textit{Physics} (7) \\
GPT L-Anx Psy & 53 & 110 & 0.04 & 1.92 & \textit{Science} (52), \textit{Scientist} (22), Try it (10) \\
\bottomrule
\end{tabular}
\end{table}

\begin{table}[!hbpt]
\scriptsize
\centering
\caption{Network features for the semantic frame of \textit{\textbf{Research}}.}
\label{tab:network_features_research}
\begin{tabular}{lccccp{10cm}}
\toprule
Sample & $N_v$ & $N_e$ & $C_i$ & $l_G$ & Hubs (top 5\% of degree distribution) \\
\midrule
Experts & 15 & 24 & 0.11 & 1.77 & Research (14) \\
H-S South & 13 & 25 & 0.20 & 1.68 & Research (12) \\
H-Anx H-S & 65 & 168 & 0.05 & 1.92 & \textit{Research} (64), \textit{Discovery} (23), \textit{Scientist} (15), \textit{Laboratory} (13) \\
L-Anx H-S & 60 & 189 & 0.08 & 1.89 & \textit{Research} (59), \textit{Science} (20), \textit{Laboratory} (17) \\
GPT H-Anx H-S & 11 & 13 & 0.07 & 1.76 & \textit{Research} (10) \\
GPT L-Anx H-S & 12 & 14 & 0.05 & 1.79 & \textit{Research} (11) \\
Physics & 27 & 55 & 0.09 & 1.84 & \textit{Research} (26), \textit{Science} (11) \\
GPT Physics & 7 & 6 & 0.00 & 1.71 & \textit{Research} (6) \\
H-Anx Psy & 40 & 208 & 0.23 & 1.73 & \textit{Research} (39), \textit{Science} (22) \\
L-Anx Psy & 52 & 245 & 0.15 & 1.82 & \textit{Research} (51), \textit{Science} (25), \textit{Curiosity} (22), \textit{Study} (22) \\
GPT H-Anx Psy & 16 & 31 & 0.15 & 1.74 & \textit{Research} (15) \\
GPT L-Anx Psy & 18 & 33 & 0.12 & 1.78 & \textit{Research} (17) \\
\bottomrule
\end{tabular}
\end{table}

\begin{table}[!hbpt]
\scriptsize
\centering
\caption{Network features for the semantic frame of \textit{\textbf{Professor}}.}
\label{tab:network_features_professor}
\begin{tabular}{lccccp{10cm}}
\toprule
Sample & $N_v$ & $N_e$ & $C_i$ & $l_G$ & Hubs (top 5\% of degree distribution) \\
\midrule
Experts & 4 & 3 & 0.00 & 1.50 & Professor (3) \\
H-S South & 17 & 30 & 0.12 & 1.78 & Professor (16) \\
H-Anx H-S & 3 & 2 & 0.00 & 1.33 & Professor (2) \\
L-Anx H-S & 2 & 1 & 0.00 & 1.00 & Professor (1), \textit{Teacher} (1) \\
GPT H-Anx H-S & 17 & 22 & 0.05 & 1.84 & \textit{Professor} (16) \\
GPT L-Anx H-S & 14 & 15 & 0.03 & 1.84 & \textit{Professor} (13) \\
GPT Physics & 9 & 9 & 0.04 & 1.75 & \textit{Professor} (8) \\
H-Anx Psy & 171 & 610 & 0.03 & 1.96 & \textit{Professor} (170), \textit{Teacher} (81), \textit{School} (41), \textit{Mathematics} (32), \textit{Exam} (29), \textit{Anxiety} (28), Work (27), \textit{University} (27), \textit{Psychology} (26), \textit{Knowledge} (26) \\
L-Anx Psy & 172 & 729 & 0.04 & 1.95 & \textit{Professor} (171), \textit{Teacher} (90), \textit{School} (51), \textit{University} (45), Work (36), \textit{Exam} (35), \textit{Mathematics} (35), \textit{Student} (32), \textit{Study} (30) \\
GPT H-Anx Psy & 33 & 57 & 0.05 & 1.89 & \textit{Professor} (32), \textit{Teacher} (17) \\
GPT L-Anx Psy & 35 & 48 & 0.02 & 1.92 & \textit{Professor} (34), \textit{Teacher} (11) \\
\bottomrule
\end{tabular}
\end{table}

\begin{table}[!hbpt]
\scriptsize
\centering
\caption{Network features for the semantic frame of \textit{\textbf{Teacher}}.}
\label{tab:network_features_teacher}
\begin{tabular}{lccccp{10cm}}
\toprule
Sample & $N_v$ & $N_e$ & $C_i$ & $l_G$ & Hubs (top 5\% of degree distribution) \\
\midrule
Experts & 16 & 27 & 0.11 & 1.77 & Teacher (15) \\
H-S South & 62 & 159 & 0.05 & 1.92 & Teacher (61), \textit{School} (23), \textit{Student} (16), Classroom (12) \\
H-Anx H-S & 2 & 1 & 0.00 & 1.00 & \textit{Teacher} (1), \textit{Teacher} (1) \\
L-Anx H-S & 2 & 1 & 0.00 & 1.00 & \textit{Teacher} (1), \textit{Teacher} (1) \\
GPT H-Anx H-S & 32 & 37 & 0.01 & 1.93 & \textit{Teacher} (31), \textit{Professor} (6) \\
GPT L-Anx H-S & 36 & 40 & 0.01 & 1.94 & \textit{Teacher} (35), \textit{Knowledge} (3), \textit{Curiosity} (3), \textit{Professor} (3) \\
Physics & 2 & 1 & 0.00 & 1.00 & \textit{Teacher} (1), \textit{Teacher} (1) \\
GPT Physics & 17 & 17 & 0.01 & 1.88 & \textit{Teacher} (16) \\
H-Anx Psy & 197 & 667 & 0.02 & 1.97 & \textit{Teacher} (196), \textit{Professor} (81), \textit{School} (47), \textit{Mathematics} (32), \textit{Task} (31), \textit{University} (31), \textit{Knowledge} (30), Work (29), \textit{Anxiety} (29), \textit{Exam} (26) \\
L-Anx Psy & 203 & 628 & 0.02 & 1.97 & \textit{Teacher} (202), \textit{Professor} (90), \textit{School} (52), \textit{University} (38), \textit{Exam} (33), \textit{Knowledge} (32), Work (31), \textit{Mathematics} (28), Degree (25), \textit{Know} (22), \textit{Study} (20) \\
GPT H-Anx Psy & 65 & 93 & 0.01 & 1.96 & \textit{Teacher} (64), \textit{Professor} (17), \textit{Development} (8), \textit{Psychology} (6) \\
GPT L-Anx Psy & 69 & 86 & 0.01 & 1.96 & \textit{Teacher} (68), \textit{Professor} (11), \textit{Development} (5), \textit{Curiosity} (4) \\
\bottomrule
\end{tabular}
\end{table}

\begin{table}[!hbpt]
\scriptsize
\centering
\caption{Network features for the semantic frame of \textit{\textbf{Scientist (male)}}.}
\label{tab:network_features_scientist (male)}
\begin{tabular}{lccccp{10cm}}
\toprule
Sample & $N_v$ & $N_e$ & $C_i$ & $l_G$ & Hubs (top 5\% of degree distribution) \\
\midrule
Experts & 10 & 9 & 0.00 & 1.80 & Scientist (9) \\
H-S South & 32 & 98 & 0.14 & 1.80 & Scientist (31), Study (13) \\
H-Anx H-S & 43 & 130 & 0.10 & 1.86 & \textit{Scientist} (42), \textit{Laboratory} (17), \textit{Scientist} (17) \\
L-Anx H-S & 46 & 105 & 0.06 & 1.90 & \textit{Scientist} (45), \textit{Science} (15), \textit{Laboratory} (14), \textit{Experiment} (14) \\
GPT H-Anx H-S & 28 & 30 & 0.01 & 1.92 & \textit{Scientist} (27), \textit{Curiosity} (4) \\
GPT L-Anx H-S & 26 & 30 & 0.02 & 1.91 & \textit{Scientist} (25), \textit{University} (4) \\
Physics & 26 & 41 & 0.05 & 1.87 & \textit{Scientist} (25), \textit{Laboratory} (8) \\
GPT Physics & 12 & 12 & 0.02 & 1.82 & \textit{Scientist} (11) \\
H-Anx Psy & 66 & 157 & 0.04 & 1.93 & \textit{Scientist} (65), \textit{Science} (21), \textit{Curiosity} (14), \textit{Physics} (13) \\
L-Anx Psy & 60 & 198 & 0.08 & 1.89 & \textit{Scientist} (59), \textit{Science} (26), \textit{Physics} (21) \\
GPT H-Anx Psy & 57 & 76 & 0.01 & 1.95 & \textit{Scientist} (56), \textit{Psychology} (11), \textit{University} (8) \\
GPT L-Anx Psy & 71 & 105 & 0.01 & 1.96 & \textit{Scientist} (70), \textit{Science} (22), \textit{University} (9), \textit{Curiosity} (6) \\
\bottomrule
\end{tabular}
\end{table}

\begin{table}[!hbpt]
\scriptsize
\centering
\caption{Network features for the semantic frame of \textit{\textbf{Scientist (female)}}.}
\label{tab:network_features_scientist (female)}
\begin{tabular}{lccccp{10cm}}
\toprule
Sample & $N_v$ & $N_e$ & $C_i$ & $l_G$ & Hubs (top 5\% of degree distribution) \\
\midrule
Experts & 10 & 9 & 0.00 & 1.80 & Scientist (9) \\
H-Anx H-S & 37 & 96 & 0.10 & 1.86 & \textit{Scientist} (36), \textit{Scientist} (17) \\
L-Anx H-S & 45 & 111 & 0.07 & 1.89 & \textit{Scientist} (44), \textit{Science} (17), \textit{Research} (16) \\
Physics & 29 & 40 & 0.03 & 1.90 & \textit{Scientist} (28), \textit{Laboratory} (6), \textit{Research} (6) \\
H-Anx Psy & 18 & 31 & 0.10 & 1.80 & \textit{Scientist} (17) \\
L-Anx Psy & 24 & 38 & 0.06 & 1.86 & \textit{Scientist} (23), \textit{Science} (8) \\
\bottomrule
\end{tabular}
\end{table}

\begin{table}[!hbpt]
\scriptsize
\centering
\caption{Network features for the semantic frame of \textit{\textbf{School}}.}
\label{tab:network_features_school}
\begin{tabular}{lccccp{10cm}}
\toprule
Sample & $N_v$ & $N_e$ & $C_i$ & $l_G$ & Hubs (top 5\% of degree distribution) \\
\midrule
Experts & 52 & 101 & 0.04 & 1.92 & \textit{School} (51), \textit{University} (14), \textit{Science} (9) \\
H-S South & 78 & 255 & 0.06 & 1.92 & \textit{School} (77), Teacher (23), \textit{Mathematics} (21), \textit{Physics} (18) \\
H-Anx H-S & 75 & 219 & 0.05 & 1.92 & \textit{School} (74), \textit{Anxiety} (19), \textit{Stress} (14), \textit{Mathematics} (14) \\
L-Anx H-S & 62 & 128 & 0.04 & 1.93 & \textit{School} (61), \textit{Mathematics} (13), \textit{Research} (10), \textit{Teacher} (10) \\
GPT H-Anx H-S & 3 & 2 & 0.00 & 1.33 & \textit{School} (2) \\
GPT L-Anx H-S & 2 & 1 & 0.00 & 1.00 & \textit{Average} (1), \textit{School} (1) \\
Physics & 26 & 27 & 0.01 & 1.92 & \textit{School} (25), \textit{Stress} (2), Comrades (2), Study (2), Friends (2) \\
H-Anx Psy & 164 & 888 & 0.05 & 1.93 & \textit{School} (163), \textit{University} (51), \textit{Anxiety} (50), \textit{Teacher} (47), \textit{Mathematics} (44), \textit{Exam} (43), Work (43), \textit{Professor} (41), \textit{Challenge} (38), \textit{Boredom} (38) \\
L-Anx Psy & 161 & 865 & 0.06 & 1.93 & \textit{School} (160), \textit{University} (61), \textit{Teacher} (52), \textit{Professor} (51), \textit{Exam} (41), Work (41), \textit{Mathematics} (36), \textit{Anxiety} (36), \textit{Boredom} (33) \\
GPT H-Anx Psy & 3 & 3 & 1.00 & 1.00 & Tests (2), \textit{School} (2), \textit{University} (2) \\
GPT L-Anx Psy & 4 & 3 & 0.00 & 1.50 & \textit{School} (3) \\
\bottomrule
\end{tabular}
\end{table}

\begin{table}[!hbpt]
\scriptsize
\centering
\caption{Network features for the semantic frame of \textit{\textbf{University}}.}
\label{tab:network_features_university}
\begin{tabular}{lccccp{10cm}}
\toprule
Sample & $N_v$ & $N_e$ & $C_i$ & $l_G$ & Hubs (top 5\% of degree distribution) \\
\midrule
Experts & 48 & 93 & 0.04 & 1.92 & \textit{University} (47), \textit{School} (14), \textit{Science} (9) \\
H-S South & 36 & 110 & 0.13 & 1.83 & \textit{University} (35), Study (15) \\
H-Anx H-S & 8 & 14 & 0.33 & 1.50 & University (7) \\
L-Anx H-S & 10 & 23 & 0.39 & 1.49 & University (9) \\
GPT H-Anx H-S & 25 & 35 & 0.04 & 1.88 & \textit{University} (24), Degree (8) \\
GPT L-Anx H-S & 20 & 29 & 0.06 & 1.85 & \textit{University} (19) \\
Physics & 5 & 5 & 0.17 & 1.50 & University (4) \\
GPT Physics & 11 & 10 & 0.00 & 1.82 & \textit{University} (10) \\
H-Anx Psy & 155 & 863 & 0.06 & 1.93 & \textit{University} (154), Work (51), \textit{School} (51), \textit{Exam} (47), Degree (44), \textit{Challenge} (40), \textit{Anxiety} (40), \textit{Psychology} (39) \\
L-Anx Psy & 163 & 1026 & 0.07 & 1.92 & \textit{University} (162), \textit{School} (61), Work (60), \textit{Exam} (55), Degree (47), \textit{Professor} (45), \textit{Anxiety} (44), \textit{Future} (40), \textit{Student} (39) \\
GPT H-Anx Psy & 51 & 99 & 0.04 & 1.92 & \textit{University} (50), \textit{Psychology} (9), \textit{Scientist} (8), \textit{Exam} (8), \textit{Development} (8) \\
GPT L-Anx Psy & 53 & 75 & 0.02 & 1.95 & \textit{University} (52), \textit{Scientist} (9), \textit{Science} (7) \\
\bottomrule
\end{tabular}
\end{table}

\FloatBarrier

\subsection{Cue words} \label{cue_words}
Tables \ref{tab:cue_words_summary} and \ref{tab:cue_words_admin} report respectively all cue-word sets and their administration in the free-association task across samples/subsamples. Sets differed due to practical constraints across data-collection waves, but all targeted the same domains of interest.

\begin{table}[!hbpt]
\centering
\caption{Cue-word sets and their administration across samples/subsamples. Subgroup size is notated as $n$.}
\label{tab:cue_words_admin}
\begin{tabular}{lll}
\toprule
Set & $N_w$ & Administered to (human / GPT) \\
\midrule
1 & 50 & Experts; High Schoolers (South).\\
  &    & 10 core cues in fixed order (italic in Table~\ref{tab:cue_words_summary}) + 40 random cues\\
  &    & drawn per participant from a 390-word Wikipedia-derived pool (Complex Systems/Physics/Math/Bio/Chem/Psych).\\
2 & 51 & Psychology undergraduates subsample ($n=57$).\\
3 & 40 & Psychology undergraduates subsample ($n=70$); all GPT-oss simulated students.\\
4 & 41 & Psychology undergraduates subsample ($n=166$).\\
5 & 42 & High Schoolers (North); Physics undergraduates; Psychology undergraduates subsample ($n=23$).\\
\bottomrule
\end{tabular}
\end{table}

\begingroup
\setlength{\extrarowheight}{0pt}
\setlength{\aboverulesep}{0pt}
\setlength{\belowrulesep}{0pt}
\setlength{\extrarowheight}{0pt}
\begin{table}[!hbpt]
\footnotesize
\centering
\caption{All the cue words sets used for the association task with all our samples (or sub-samples). $N_w$ stands for the number of words present in each set.}
\label{tab:cue_words_summary}
{\renewcommand{\arraystretch}{0.1}}
\begin{tabular}{m{0.5cm}|m{3cm}|m{3.1cm}|m{2.3cm}|m{3.2cm}|m{2.8cm}}
\toprule
Set & STEM Disciplines & Education and Research & Evaluation & Learning and Motivation & Mental Health \\
\midrule
\rotatebox{90}{1 ($N_w = 50$)} & Science, \textit{Physics}, \textit{Mathematics}, \textit{Chemistry}, \textit{Biology}, Statistics, Electricity, Computer, Evolution, Dna, Experiment, Cell, Derivative, Space, \textit{System}, Graph, Theorem, Trigonometry, Integral, Model, \textit{Complex}, Chaos, Solid, Number, Hub, Network, Theory & \textit{University}, \textit{School}, Conference, Lab, Studies, Teaching, Student, Degree, Mathematician, Prisoner & Analysis, Game, Word, Loop, Market & Language, Person & \textit{Life}, Brain, Water, Scientific, Cloud, \textit{Art} \\ \midrule
\rotatebox{90}{2 ($N_w = 51$)}
 & Algorithm, Biology, Calculation, Demonstration, Equation, Physics, Engineering, Mathematics, Mathematician, Numbers, Performance in mathematics, Statistics, Technology, Math test, Utility of mathematics & Class, Parent, Teacher, Student, Board, Concept map, Notes & Exam & Change of plans, Task, Concentration, Confusion, Curiosity, Distraction, Errors, Set a goal, I commit, Importance, Intelligence, Store, Motivation, Plan, Problem, Plan, Program, Review, Postpone, Expiration, Challenge, Effort, Solution, Study, Success, Keep track & Anxiety, Failure \\ \midrule
\rotatebox{90}{3 ($N_w = 40$)}
 & Biology, Equation, Physics, Informatics, Mathematics, Statistics, Science, Scientist, Innovation, Neuroscience & Teacher, Professor, University & Exam, Trial, Test, Average, Assessment & Creativity, Curiosity, Fun, Challenge, Model, Know, Development, Passion & Anxiety, Attachment, Well being, Cognition, Emotion, Mind, Personality, Psychology, Psychopathology, Psychotherapy, Therapist \\ \midrule
\rotatebox{90}{4 ($N_w = 41$)}
 & Biology, Equation, Physics, Mathematics, Numbers, Statistics, Technology, Graphic, Science, Discovery, Neuroscience & Teacher, Student, Professor, School, University & Exam, Exercise, Grade & Task, Knowledge, Creativity, Curiosity, Challenge, Future, Model, Relation & Anxiety, Well being, Cognition, Emotion, Frustration, Mind, Boredom, Personality, Psychology, Therapy, Stress \\ \midrule
\rotatebox{90}{5 ($N_w=42$)}
 & Analyses, Equation, Physics, Informatics, Mathematics, Numbers, Statistics, Data, Hypothesis, Laboratory, Logic, Research, Science, Scientist, Scientist, Discovery, Experiment, Innovation & Companions, Teacher, Parents, School & Exam, Grade & Knowledge, Creativity, Problem, Programming, Challenge, Future, Rules & Anxiety, Art, Well being, Counseling, Failure, Embarrassment, Meditation, Panic, Fear, Stress \\
\bottomrule
\end{tabular}
\end{table}
\endgroup

\FloatBarrier

\section{Supplementary semantic frames} \label{appendix: semantic_frames}
To be thorough, and due to spatial constraints, we report here additional semantic frames obtained in our experiments.


\begin{figure}[!hbpt]
    \centering
    \includegraphics[width=0.65\linewidth]{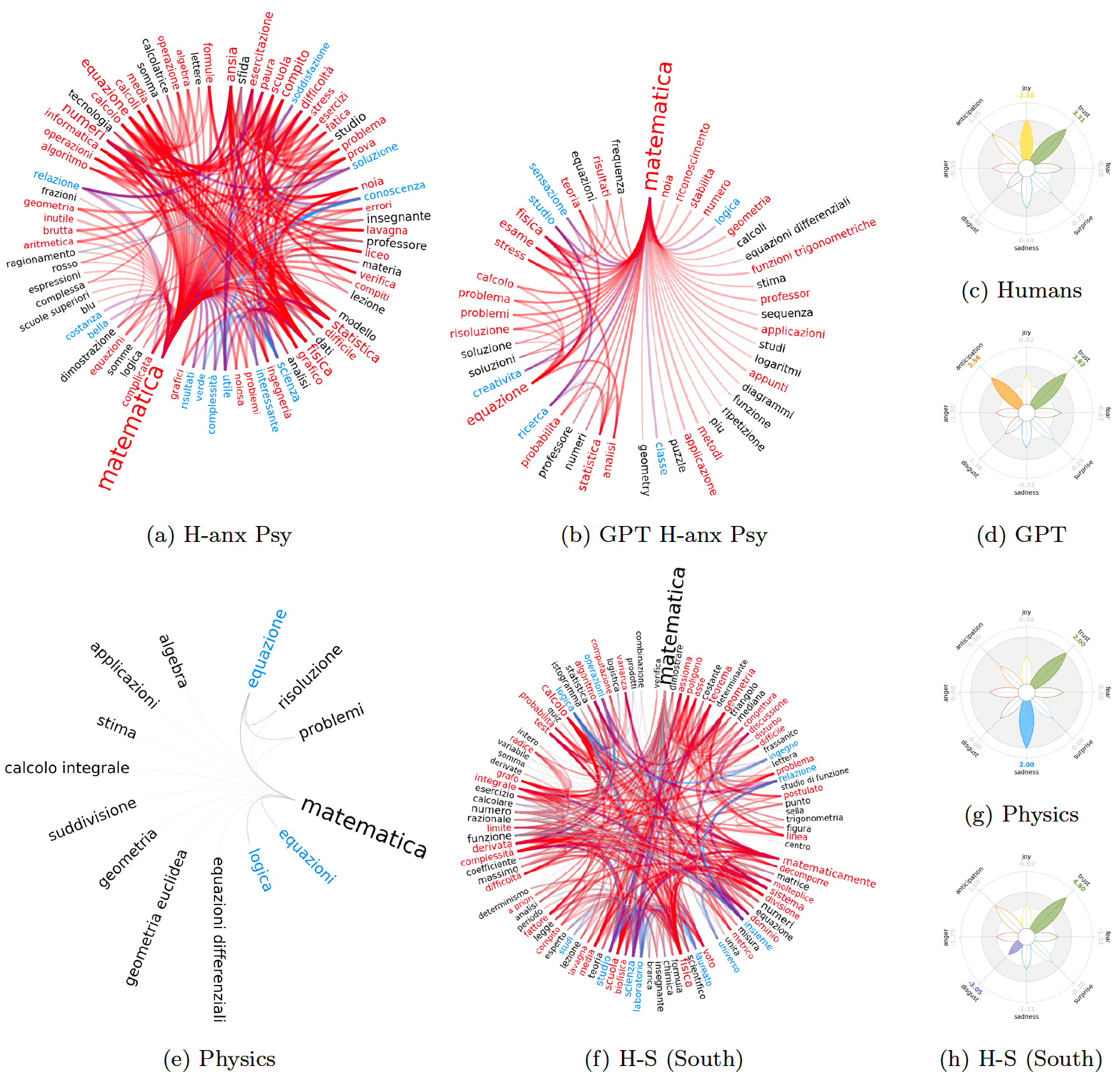}
    \caption{Semantic frames of the node \textbf{\textit{Math}}.}
    \label{fig: appen_math_semantic_frames}
\end{figure}


\begin{figure}[!hbpt]
    \centering
    \includegraphics[width=0.65\linewidth]{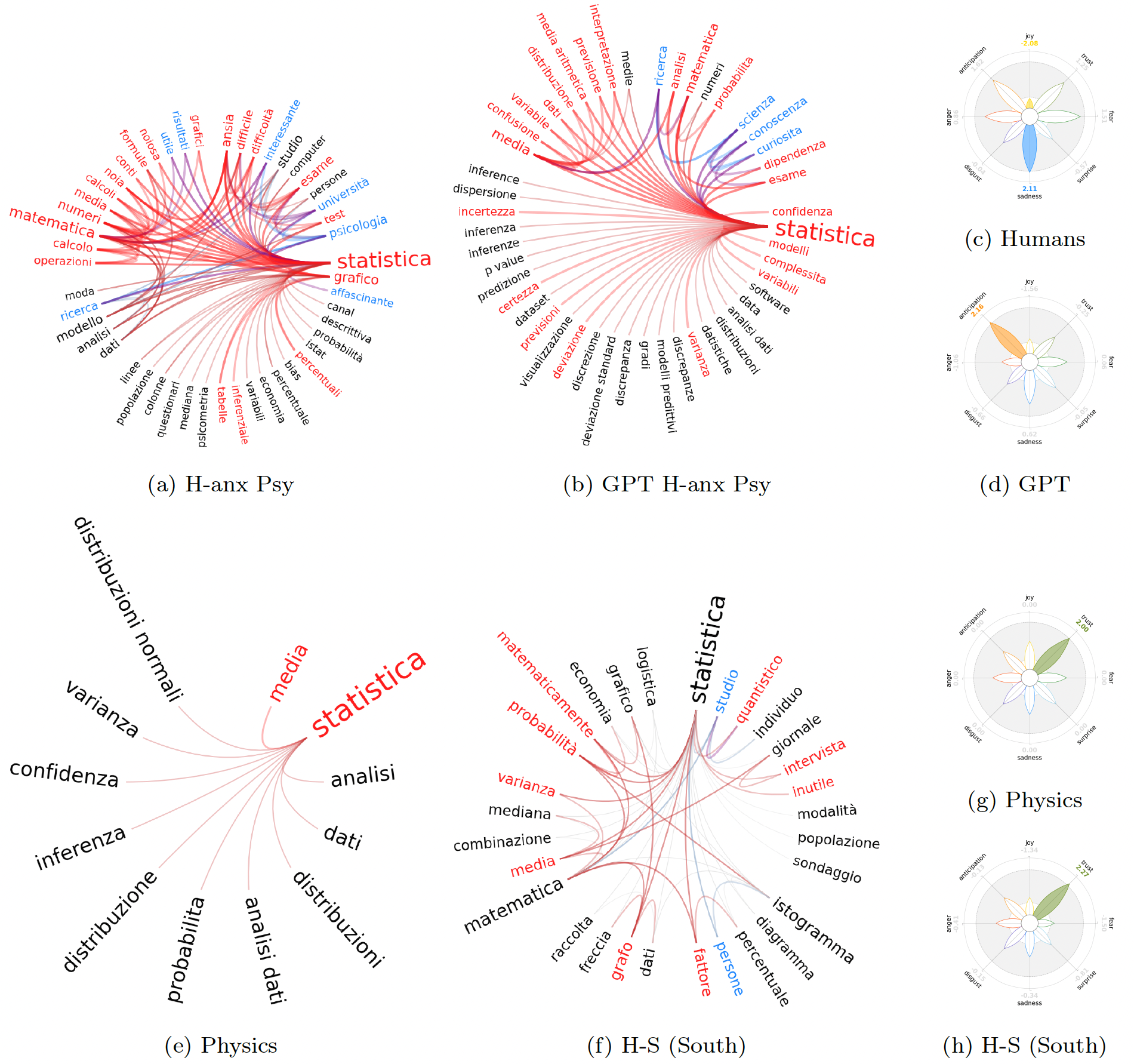}
    \caption{Semantic frames of the node \textbf{\textit{Statistics}}.}
    \label{fig: appen_statistics_semantic_frames}
\end{figure}


\begin{figure}[!hbpt]
    \centering
    \includegraphics[width=0.65\linewidth]{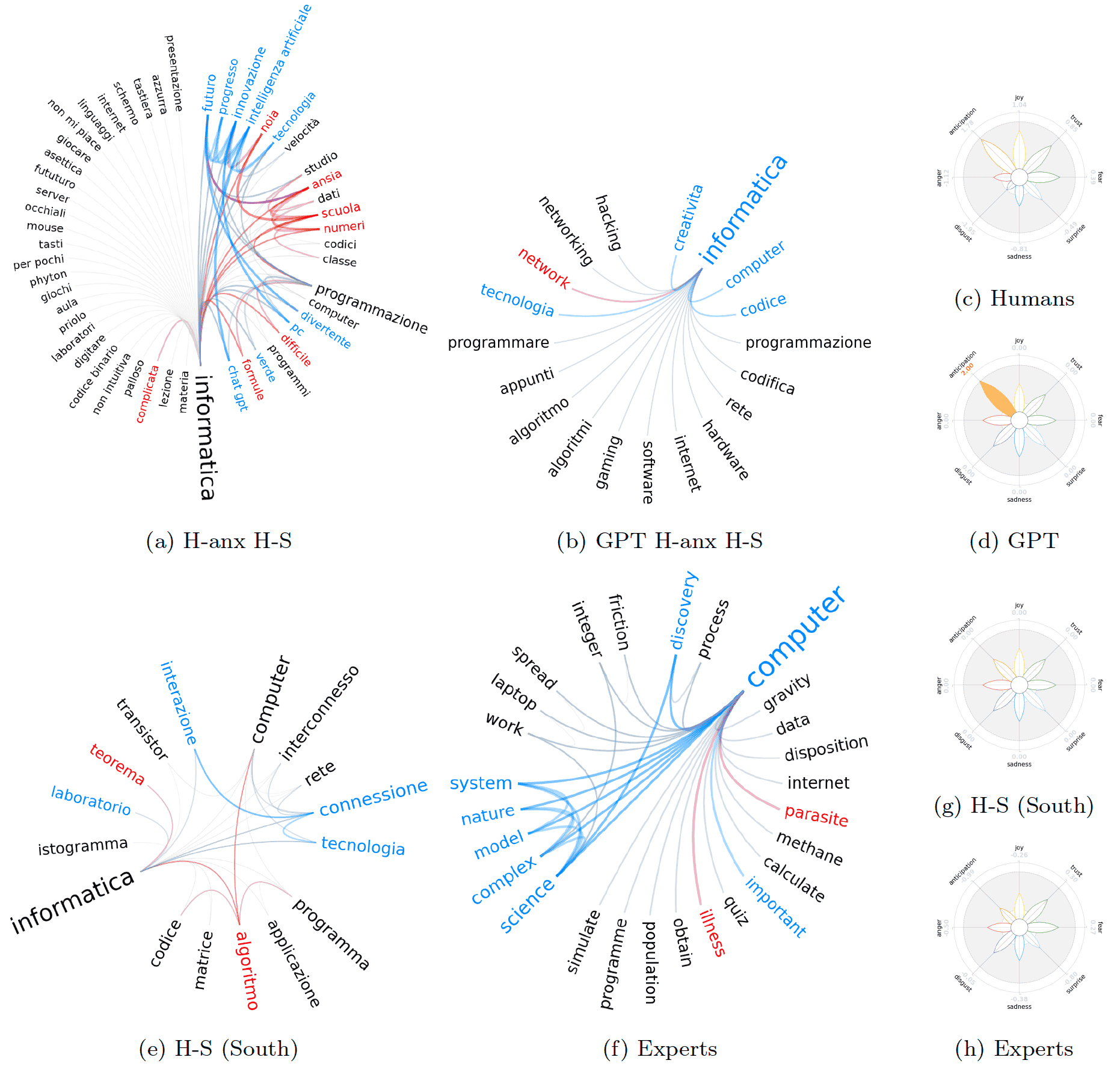}
    \caption{Semantic frames of the node \textbf{\textit{Computer Science}}.}
    \label{fig: appen_informatics_semantic_frames}
\end{figure}


\begin{figure}[!hbpt]
    \centering
    \includegraphics[width=0.65\linewidth]{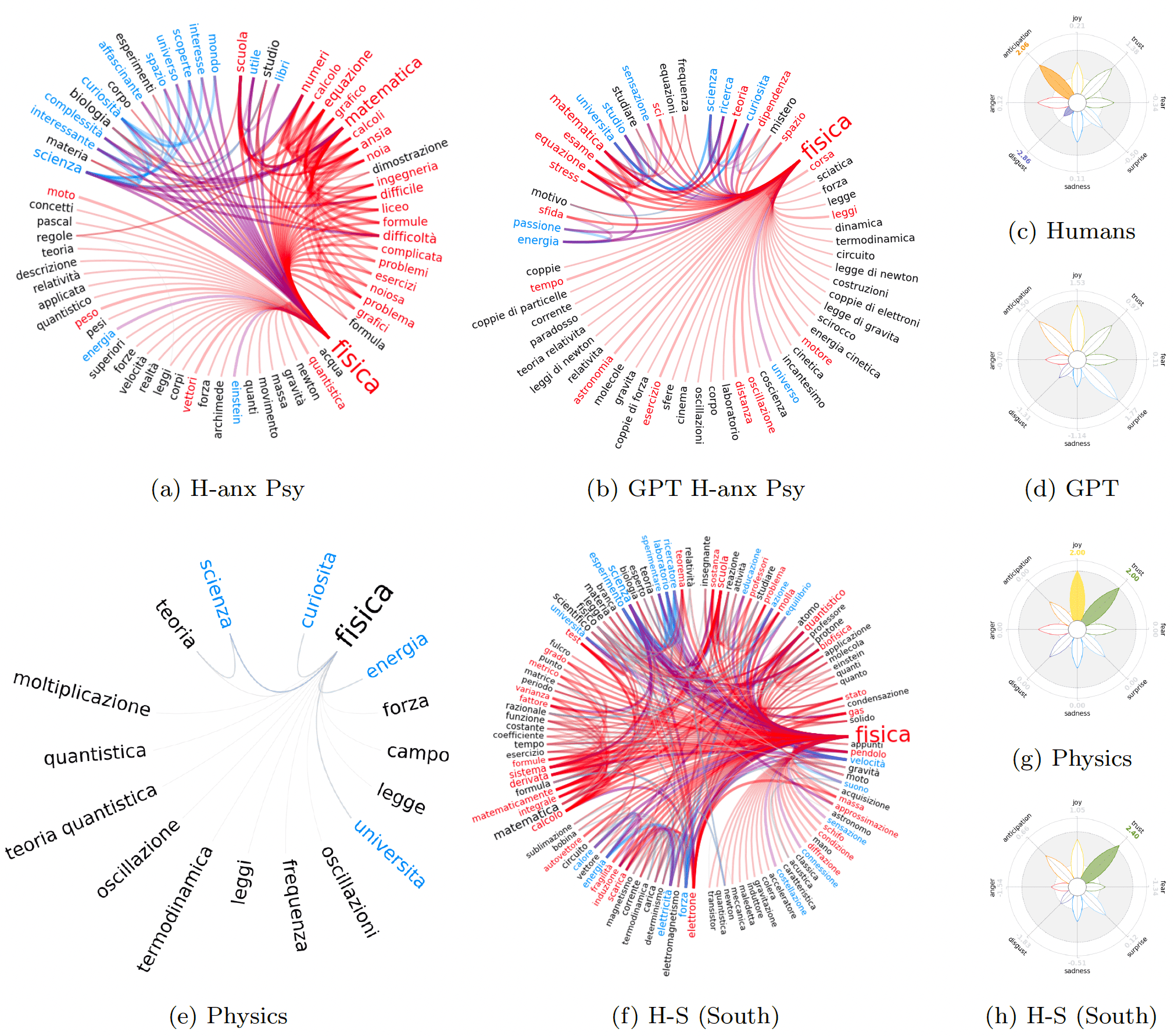}
    \caption{Semantic frames of the node \textbf{\textit{Physics}}.}
    \label{fig: appen_physics_semantic_frames}
\end{figure}


\begin{figure}[!hbpt]
    \centering
    \includegraphics[width=0.65\linewidth]{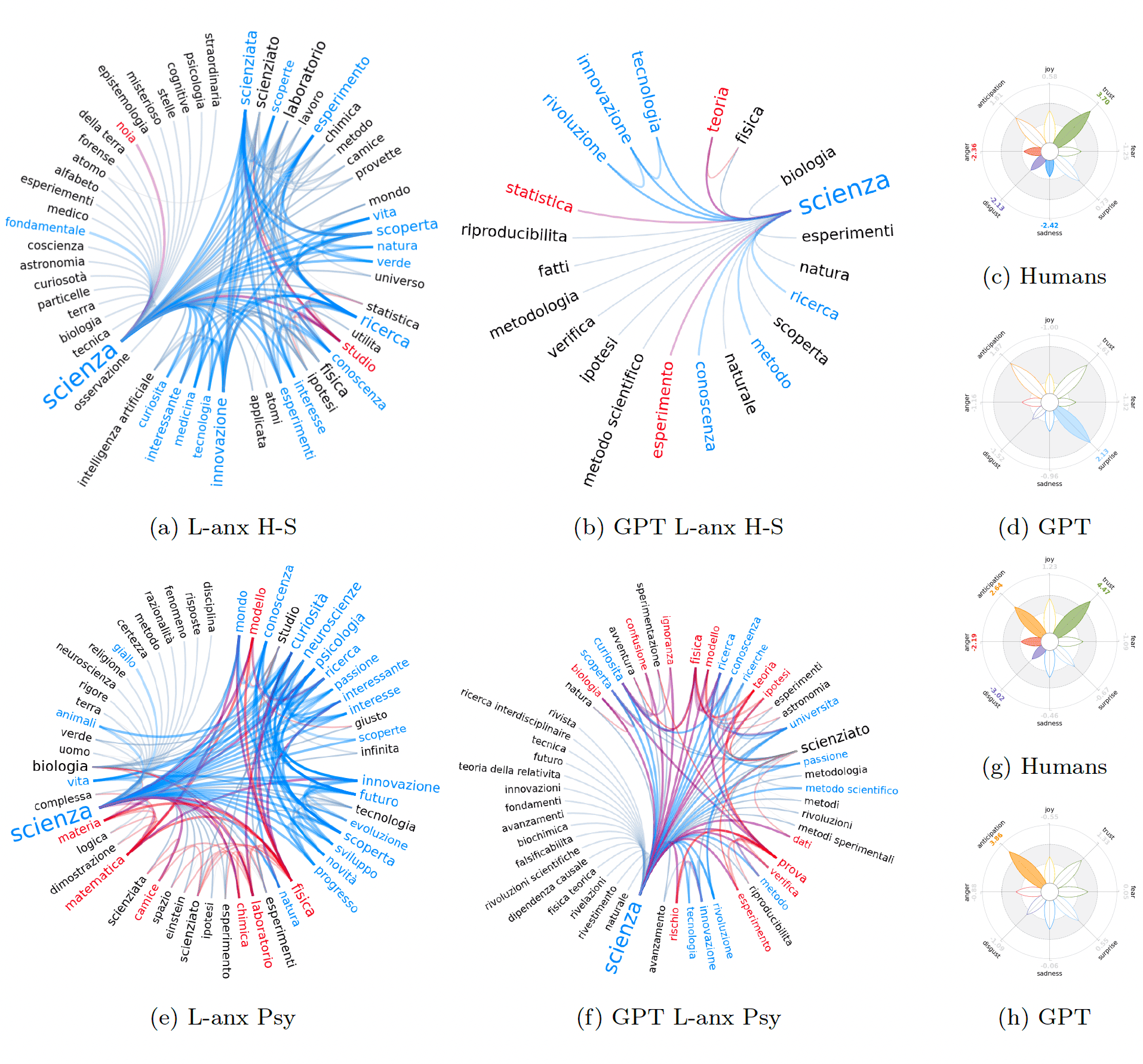}
    \caption{Semantic frames of the node \textbf{\textit{Science}}.}
    \label{fig: appen_science_semantic_frames}
\end{figure}

\section{Methods: supplementary materials}\label{appendix:methods}
\subsection{Mathematical notation found in the concreteness ratings tables} \label{appendix:concreteness_notation}
Here is an explanation of each variable found in the tables of the concreteness ratings:
\begin{itemize}
    \item \textbf{Group}: The specific \textit{experimental group or subgroup} being analysed;
    \item \textbf{Keyword}: The \textit{concept} or word whose associated empirical frame concreteness is being rated;
    \item \textbf{k}: The \textit{degree} of a specific semantic frame;
    \item $\mathbf{\bar{x} - \hat{\mu}_0}$: The \textit{difference in mean concreteness scores} between the empirical frames and the baseline random sample;
    \item$\mathbf{Cohen's \ d}$: A parametric measure of the \textit{effect size} for the difference in means \citep{cohenStatisticalPowerAnalysis1988};
    \item$\mathbf{Z}$: The \textit{z-score} used for the test of difference in means;
    \item$\mathbf{\bar{x}}$: The \textit{mean concreteness score} of the empirical frames;
    \item$\mathbf{Cliff's \ \delta}$: A non-parametric measure of the \textit{effect size} for the difference in means;
\end{itemize}

\subsection{Prompt}\label{prompt}
English translation of the prompt used to impersonate human counterparts:
\begin{quote}
    \textit{You are a} \textit{student}\texttt{\{gender\}} \textit{of Italian nationality}, \textit{aged} \texttt{\{age\}} \textit{. You are enrolled in the} \texttt{\{year\}} \textit{year of} \texttt{\{education\}}\textit{. You grew up and live in} \texttt{\{socioeconomic\}} \textit{socio-economic conditions. Therefore, remember that the responses you provide in the task should be original, creative, and consistent with your unique characteristics.}.
\end{quote}

\end{document}